\pdfoutput=1
\documentclass[11pt]{article}
\usepackage{times}
\usepackage{latexsym}
\usepackage[T1]{fontenc}
\usepackage[utf8]{inputenc}
\usepackage{microtype}

\usepackage{acl}

\usepackage{float}
\usepackage{wrapfig}
\usepackage{subfig}
\usepackage{footnotehyper}

\usepackage[main=english]{babel}
\usepackage[autostyle]{csquotes}
\usepackage{microtype}
\usepackage{enumitem}
\usepackage{lipsum}

\usepackage{tabularx}
\usepackage{booktabs}
\usepackage{multirow}
\usepackage{makecell}

\makeatletter
\newcommand{\customlabel}[2]{%
 \@bsphack\begingroup
 \def\@currentlabel{#2}%
 \label{#1}%
 \endgroup\@esphack
}
\makeatother
\newcommand{\tablelabel}[1]{\customlabel{#1}{Table \arabic{table}}}
\newcommand{\figurelabel}[1]{\customlabel{#1}{Figure \arabic{figure}}}
\newcommand{\subfigurelabel}[1]{\customlabel{#1}{Figure \arabic{figure}\alph{subfigure}}}
\newcommand{\sectionlabel}[1]{\customlabel{#1}{Section \arabic{section}}}
\newcommand{\subsectionlabel}[1]{\customlabel{#1}{Section \arabic{section}.\arabic{subsection}}}

\usepackage{amsmath}
\usepackage{amsfonts}

\usepackage[size=tiny]{todonotes}

\newcommand{\NLP}[0]{\texttt{NLP}}
\newcommand{\ML}[0]{\texttt{ML}}
\newcommand{\Hist}[0]{\texttt{Hist}}
\newcommand{\ttest}[0]{\mbox{\textit{t}-test}}
\newcommand{\Htest}[0]{\mbox{\textit{H} test}}
\newcommand{\positive}[1]{\textcolor{blue}{#1}}
\newcommand{\negative}[1]{\textcolor{red}{#1}}

\title{Did AI get more negative recently?}

\author{
    Dominik Beese$^1$ \and Beg{\"u}m Altunba{\c{s}}$^2$ \and G{\"o}rkem G{\"u}zeler$^2$ \and Steffen Eger$^3$ \\
    \texttt{dominik.beese@stud.tu-darmstadt.de,steffen.eger@uni-bielefeld.de} \\
    \\
    $^1$ Technische Universität Darmstadt \quad $^2$ Technische Universität München \\
    $^3$ Natural Language Learning Group (NLLG), Faculty of Technology, Bielefeld University \\
}

\begin{document}

\maketitle
\begin{abstract}
In this paper, we classify scientific articles in the domain of natural language processing (NLP) and machine learning (ML), as core subfields of artificial intelligence (AI), into whether (i) they extend the current state-of-the-art by the introduction of novel techniques which beat existing models or whether (ii) they mainly criticize the existing state-of-the-art, i.e. that it is deficient with respect to some property (e.g. wrong evaluation, wrong datasets, misleading task specification). We refer to contributions under (i) as having a `positive stance' and contributions under (ii) as having a `negative stance' (to related work). We annotate over 1.5 k papers from NLP and ML to train a SciBERT-based model to automatically predict the stance of a paper based on its title and abstract. We then analyse large-scale trends on over 41 k papers from the last approximately 35 years in NLP and ML, finding that papers have become substantially more positive over time, but negative papers also got more negative and we observe considerably more negative papers in recent years. Negative papers are also more influential in terms of citations they receive.
\end{abstract}

\section{Introduction}\sectionlabel{sec:introduction}

Deep learning has revolutionized machine learning (ML) and natural language processing (NLP) in the last decade. In particular, deep learning has led to unprecedented performance gains on a large number of NLP and ML tasks, including machine translation \citep{lample2018unsupervised}, image classification \citep{krizhevsky2012imagenet}, natural language understanding \citep{devlin2018bert}, and text generation \citep{radford2019language}.

\begin{table}[!hbtp]
    \centering
    \begin{tabularx}{8cm}{X}
        \toprule
        \textbf{Abstract} (excerpt) \\
        \midrule
        \enquote{We are surprised to find that BERT’s peak performance of 77\% on the Argument Reasoning Comprehension Task reaches just three points below the average untrained human baseline. \negative{However, we show that this result is entirely accounted for by exploitation of spurious statistical cues in the dataset. We analyze the nature of these cues and demonstrate that a range of models all exploit them. This analysis informs the construction of an adversarial dataset on which all models achieve random accuracy.} [...]} \\
        \bottomrule
    \end{tabularx}
    \caption{Example of a paper with negative stance. The underlying paper is \citet{niven-kao-2019-probing}. Negative elements are highlighted by us in red.}
    \tablelabel{tab:example}
\end{table}

On the other hand, there has seemingly also been a recent surge of papers highlighting limitations of (deep learning) approaches, including claims about models exploiting dataset biases \citep{niven-kao-2019-probing}, flawed evaluation \citep{marie-etal-2021-scientific}, and general `troubling trends' in ML practice \citep{Lipton:2019}.

Indeed, from a historic perspective, deep learning -- formerly known under the name of `artificial neural networks' -- is a prime exemplar of a technology that has received very mixed assessments over time, ranging from initial hype to negative and positive appraisal in repeating cycles \citep{Hendler:2008,guo2020application}.

Motivated especially by (recent) observations of negative assessment of individual papers regarding the existing literature and its claims \citep{bowman-2022-dangers}, we define a new NLP task of determining the \emph{stance} of a paper (with respect to its related work).\footnote{This concept is related to positive/negative citations within a paper, which have been annotated in a few works, e.g. \citet{teufel_automatic,athar-teufel-2012-context,abu-jbara-etal-2013-purpose,catalini:2015,yousif2019survey}. Our work goes beyond individual citations and assesses the stance of the authors' main message.} We take a prototypical paper of negative stance to be one that concludes that related work is basically flawed, mistaken and potentially based on false assumptions (cf.\ \ref{tab:example}); negative papers could also be referred to as \emph{critique papers}, which disrupt current knowledge. By contrast, a prototypical paper of positive stance is one that generally accepts the premises of related work (even though it may identify specific -- minor -- issues and weaknesses), extends it and sets a new state-of-the-art. Positive papers could also be referred to as \emph{extension papers},  which build upon and extend current knowledge.\footnote{We note that there is an asymmetry between positive and negative papers in our definition: while negative papers explicitly relate to related work, positive papers especially relate to themselves, highlighting their own positive contributions. Nonetheless, we take positive papers to implicitly accept the premises of related work, thus they have an \emph{implicit} positive stance to it. We discuss more on this in \ref{sec:annotation}. Calling positive papers incremental, as one reviewer suggested, would seem like a biased judgement concerning their quality.} (Any particular paper may mix positive and negative elements, so we treat the task as a continuous regression problem.)

We hold this task important in order to be able to analyze trends in science and its evolution \citep{catalini:2015}, which could potentially anticipate pessimistic future developments (the end of the party). By comparing trends across two core  disciplines of artificial intelligence (AI) (arguably one of the most dynamic, promising and intriguing current research fields) -- ML and NLP -- we can also contrast the evolution and current state of each. The task is also timely, as interest in the analysis of scientific literature in the NLP community has been steadily on the rise recently (cf. \ref{sec:related_work}).

To our best knowledge, we are the first to tackle stance identification on the level of abstracts which, in contrast to individual citations, is (i) more efficient, (ii) less ambiguous, and (iii) focuses on the authors' core message. We are also the first to measure the evolution of two important scientific subfields (NLP and ML) with respect to stance over time and relate stance to important scientific success measures, i.e. citation counts and acceptance chances at venues. 
Our contributions:
\begin{itemize}[topsep=2pt,itemsep=-2pt,leftmargin=*]
    \item we define the new task of stance detection for scientific literature;
    \item we provide a human-annotated dataset of over 1.5 k scientific papers, labelled for their stance;
    \item we provide a large-scale trend analysis on over 41 k papers from the NLP and ML community in the past approximately 35 years;
    \item we address various trend questions including (i) whether negativity has recently increased, (ii) whether positive/negative papers are more influential, and (iii) whether positive/negative papers are more likely to be accepted.
\end{itemize}

\noindent We point out that `negativity', as is a focus of our work, plays a central role in various contexts: in the social sciences, signed social networks are  networks in which agents have positive and negative relations to each other, potentially explaining phenomena such as long-term disagreement \citep{altafini2012consensus,eger2016opinion}; in science, negative citations may (arguably) be a form of self-correction \citep{bordignon2020self} and publishing negative results may reduce waste on resources for disputed approaches \citep{mlinaric2017dealing}; in economics, the principle of \emph{creative destruction} \citep{aghion1990model} may explain the progress of capitalism.\footnote{Interestingly, the publication of our paper coincides with a new hype in the AI community surrounding the release of ChatGPT, a very high-quality conversational dialogue model.}

\section{Related work}\sectionlabel{sec:related_work}

Historically, analysis of scientific literature is the scope of the fields of \emph{scientometrics} and \emph{science-of-science}. Classical results include the relation between title length and the number of citations a paper receives \citep{letchford2015advantage} as well as quantitative laws underlying citation patterns or the number of co-authors of papers over time \citep{fortunato2018science}. \citet{sienkiewicz2016impact} relate textual properties of abstracts and titles of scientific papers to their popularity and find that the complexity of abstracts is positively correlated with citation counts, but abstract and title sentiment (measured as average valence/arousal of all words) are weakly correlated.\footnote{They use a lexicon lookup to measure valence/arousal, which cannot deal with the contextual usage of words, e.g. the distinction between `fail' and `not fail'. By contrast, we learn a model on human annotations to determine the stance of a paper and our notion of stance does also not coincide with sentiment, see \ref{sec:stance-vs-sentiment}. We finally target very different scientific venues, i.e. ML/NLP versus Web of Science.} In recent years, with the rise in quality of models and approaches, more and more NLP approaches are also devoted to the analysis of scientific literature. We list several relevant studies in the following.

\citet{DoesMyRebuttalMatter} ask how much the author response in the `rebuttal phase' of the peer review process influences the final scores of a reviewer, finding its impact to be marginal, especially compared with the scores of other reviewers. \citet{pei-jurgens-2021-measuring} study (un)certainty in science communication, finding differences among journals and team size.

\citet{PredictingTheRiseAndFall} predict whether a scientific topic will rise or fall in popularity based on how authors frame the topics in their work. They use a subset of the Web of Science\footnote{\url{https://clarivate.com/webofsciencegroup/solutions/web-of-science/}} Core Collection with papers from 1991 to 2010 and analyse abstracts by assigning scientific topics (e.g. stem cell research) as well as rhetorical roles (i.e. scientific background, methods used, results etc.) to phrases. They find that topics that are currently discussed as results and background are at their peak and tend to fall in popularity in the future, whereas topics that are mentioned as methods or conclusions tend to start to rise in popularity.

Arguably the paper most similar to ours is \citet{MeasuringTheEvolution}. They study the entire content of a scientific publication in order to predict its future impact, based on how citations are framed. They distinguish different \emph{functions} of citations: \textsc{Background} (the other work provides relevant information), \textsc{Uses} (usage of data, methods, etc.\ from the other work), and \textsc{Comparison or Contrast} (express similarities/differences to the other work). They analyse the evolution of the functions over (i) the course of a paper, (ii) different venues, and (iii) time. They show that NLP has seen considerable increase in consensus when authors started to use fewer \textsc{Comparison or Contrast} citations and simply acknowledged previous work as \textsc{Background}. The authors argue that these trends imply that NLP has become a \emph{rapid discovery science} \citep{Collins1994}, i.e. a particular shift a scientific field can undergo when it reaches a high level of consensus on its research topics, methods and technologies, and then starts to continually improve on each other's methods. Our approach differs from \citet{MeasuringTheEvolution} in several ways: e.g. we do not analyse individual citations, but directly evaluate the stance of a complete paper (as measured by its framing in the paper's abstract); most importantly, we are particularly interested in \emph{negative} stances, which as relation is absent in the scheme of \citet{MeasuringTheEvolution}.

A number of more recent papers also leverage or study individual citation context, including \citet{cohan-etal-2019-structural,jebari2021use,wright-augenstein-2021-citeworth,lauscher2021multicite}. We note that annotating individual citations is more costly than annotating the abstract of a paper, as we do. Annotating individual citations is also a complex task which requires context for disambiguation -- as neglected in most previous work \citep{lauscher2021multicite} -- and even then may involve looking up the cited work to understand the citation role. There may also be a bias against direct negative citations of \emph{individual} works, cf.\ also \citet{bordignon2020self}. Sentiment analysis (e.g. polarity) for individual citations is surveyed in \citet{yousif2019survey}. For example, \citet{abu-jbara-etal-2013-purpose} define a positive citation as one that explicitly states a strength of the target paper. A negative citation points to a weakness and descriptive citations are marked as neutral. We adopt a similar scheme, but apply it to the core message instead of individual citations; we also define positive papers differently\footnote{It would not make sense to define a paper as positive if it \emph{explicitly} lauds previous work as its main message; such papers would probably be rejected as having no additional value to the field of science.} --- i.e. they make a positive contribution in terms of proposing new techniques extending existing work (thereby implicitly accepting its premises) and setting a new state-of-the-art. \citet{catalini:2015} study the role of negative citations and find (among others) that they tend to decrease citation counts of the cited paper over time. We study the dual question: whether negative papers (in our sense) receive more citations. \citet{lamers2021meta} study disagreement in science across diverse fields, which is a related concept to that of negative citations, finding that there is highest disagreement in the social sciences and humanities, and lowest disagreement in mathematics and computer science (of which AI is a subfield).

Beyond classification, \citet{CanWeAutomate} use NLP models to automatically generate reviews for scientific papers. They conclude that their review generation model is not good enough to fully automate the reviewing process, but could still make the reviewer's job more effective. \citet{ReviewRobot} create automatic reviews for papers by defining multiple knowledge graphs, one extracted from the paper, one from the papers the paper cites, and one for background knowledge. \citet{SciBERT} train a language model (SciBERT; extending BERT) on a large multi-domain corpus of scientific publications. 

\section{Data}\sectionlabel{sec:data}

We extract our data from two sources: (i) the ACL~Anthology\footnote{\url{https://aclanthology.org/}} which contains papers and metadata for all major NLP events, and (ii) ML conferences.

\subsection{\NLP{} dataset}
From the ACL~Anthology, we extract papers from eight different NLP venues between 1984 and 2021.\footnote{We find by introspection that papers before 1984 often have different formatting and may even lack specifically delineated abstracts. For the same reason, we exclude papers from the CL journal before 1986.} For all venues, we only include papers from the main conference and exclude papers from workshops (by manually selecting the volumes) and contributions like book reviews and title indices (by filtering the titles). To extract the data, we download the provided metadata from the ACL~Anthology website in the BibTeX format that contains information of authors, title, venue and year. We then use Allen AI’s Science~Parse\footnote{\url{https://github.com/allenai/science-parse/} In some instances, that could not be removed automatically, Science~Parse included other parts of the paper. This leads to increased uncertainty especially for papers published before the year 2000.} to extract abstract information and collect citation information from Semantic~Scholar.\footnote{\url{https://www.semanticscholar.org/}} We refer to this dataset as \NLP{} in the following. \NLP{} contains more than 23 k papers in total. The distribution over the venues is shown in \ref{tab:nlp_distribution}.

\begin{table}[!htb]
    \centering
    \parbox{.45\linewidth}{
        \centering
        {\small\begin{tabular}{l|r}
             \toprule
             \multicolumn{1}{c|}{\textbf{Venue}} & \textbf{\# papers} \\
             \midrule
             ACL     & 6,818 \\
             EMNLP   & 5,013 \\
             COLING  & 4,647 \\
             NAACL   & 2,636 \\
             SemEval & 2,384 \\
             CoNLL   & 1,033 \\
             CL      & 952 \\
             TACL    & 351 \\
             \midrule
             Overall & 23,834 \\
             \bottomrule
        \end{tabular}}
        \caption{\NLP{} dataset.}
        \tablelabel{tab:nlp_distribution}
    }
    \hspace{4pt}
    \parbox{.45\linewidth}{
        \vspace{-3.65em}
        {\small\begin{tabular}{l|r}
             \toprule
             \multicolumn{1}{c|}{\textbf{Venue}} & \textbf{\# papers} \\
             \midrule
             NeurIPS & 7,535 \\
             AAAI    & 4,384 \\
             ICML    & 3,667 \\ 
             ICLR    & 2,720 \\
             \midrule
             Overall & 18,306 \\
             \bottomrule
        \end{tabular}}
        \caption{\ML{} dataset.}
        \tablelabel{tab:ml_distribution}
    }
\end{table}

\subsection{\ML{} dataset}
We download papers from the respective websites of NeurIPS\footnote{\url{https://papers.neurips.cc/}}, AAAI\footnote{\url{https://www.aaai.org/}}, ICML\footnote{\url{https://icml.cc/}}, and ICLR\footnote{\url{https://iclr.cc/}}, and then use Science~Parse to extract abstracts and Semantic~Scholar to collect citation information as above. \ML{} contains over 18 k papers between 1989 and 2021. The distribution over the venues is given in \ref{tab:ml_distribution}.

\section{Data annotation}\sectionlabel{sec:annotation}

We annotate the data from \NLP{} and \ML{} for each paper's \textbf{stance (towards related work)}, as given in the authors' \emph{framing} in a paper's abstract. In contrast to some related work, we do not annotate stance towards individual citations, but infer the authors' stance from the paper abstracts and titles as we are interested in the stance of the authors' \emph{overall core message}. Our focus on title and abstract is a deliberate choice resting on the following observations. (i) In abstracts, authors typically condense the most important information they intend to convey. (ii) This agrees with the insight that abstracts and titles are typically the only piece of the paper that the majority of readers consumes \citep{Andrade2011HowTW}. (iii) Annotating and classifying individual citations is also more costly, cognitively demanding and ambiguous, as outlined in \ref{sec:related_work}. (iv) We are also not interested in whether individual citations are positive or negative but in whether the whole paper (the core message) is framed positively or negatively.

\subsection{Definition of stance}
On a coarse-grained level, we consider three possible stances:
(i)~we define a prototypical paper of \underline{\emph{positive stance}} (towards related work) as one that directly or indirectly accepts its premises, builds upon and  extends the existing literature and achieves a new state-of-the-art; such a paper contains phrases and sentences such as `we present a new approach'; `we beat the state-of-the-art'; `we release a new dataset' (cf.\ \ref{sec:appendix} \ref{tab:llr}). 
(ii)~A prototypical paper of \underline{\emph{negative stance}} towards related work states that datasets, evaluation protocols or techniques are basically flawed; such a paper contains sentences of negative sentiment such as `techniques fail'; `approaches are limited'; `models are unstable'. 
(iii)~If a paper does not particularly fit into this categorization, e.g. because it discusses or summarizes previous work without criticizing it, then we consider a paper as expressing a \underline{\emph{neutral stance}}; survey or analysis papers typically fall into this category.

In the annotation, we relax this coarse-grained scheme and instead allow continuous numbers as stances, ranging from $-$1 (very negative stance) towards +1 (very positive stance), with 0 as neutral stance.\footnote{In practice, annotators typically choose `pseudo-discrete' annotation scores in steps of 0.1, e.g. $-$0.2, $-$0.1, 0.0, 0.1, 0.2.} Intuitively, the more negative/positive statements an abstract contains, the more negative/positive it is. The severity of statements also matters for the degree of positivity or negativity, e.g. `techniques fail completely' is more negative than `some techniques don't work properly'. Positive/neutral and negative papers are differentiated by the amount of \emph{direct criticism} they express against existing work: each element of criticism (and its severity) would decrease the positivity of a paper. Neutral and positive papers are differentiated in that the latter have clearly positive elements of advancing the state-of-the-art and providing better solutions.

We provide guidelines for the annotation in the \ref{sec:appendix} as well as several examples in \ref{tab:human_data} and \ref{tab:human_data2}. In the following, we describe the annotation process and provide statistics.

\subsection{Annotation statistics and procedure}\subsectionlabel{sec:procedure}
In total, we manually annotated 1,550 papers from \NLP{} and \ML{}. The distribution of papers over venues is given in \ref{tab:human_annotated_dataset_distribution}. In this human-annotated dataset, the ACL conference, which takes place annually since 1979, has most papers (225), followed by NeurIPS (210) and AAAI (205).

\begin{table}[!htb]
    \centering
    \parbox{.45\linewidth}{
        \vspace{-0.50em}
        {\small\begin{tabular}{l|r}
            \toprule
            \multicolumn{1}{c|}{\textbf{Venue}} & \textbf{\# papers} \\
            \midrule
            ACL     & 225 \\
            NeurIPS & 210 \\
            AAAI    & 205 \\
            ICLR    & 196 \\
            ICML    & 167 \\
            NAACL   & 127 \\
            \bottomrule
        \end{tabular}}
    }
    \hspace{1pt}
    \parbox{.45\linewidth}{
        {\small\begin{tabular}{l|r}
            \toprule
            \multicolumn{1}{c|}{\textbf{Venue}} & \textbf{\# papers} \\
            \midrule
            EMNLP   & 123 \\
            COLING  & 111 \\
            CL      &  89 \\
            SemEval &  50 \\
            CoNLL   &  47 \\
            \midrule
            Overall & 1,550 \\
            \bottomrule
        \end{tabular}}
    }
    \caption{Human-annotated dataset.}
    \tablelabel{tab:human_annotated_dataset_distribution}
\end{table}

In our human-annotated data, 1,018 papers (65.7\%) exhibit \emph{positive stance} ($\geq 0.1$), 277 papers (17.9\%) exhibit \emph{negative stance} ($\leq -0.1$), and 255 papers (16.5\%) are \emph{neutral} ($\in (-0.1,+0.1)$).\footnote{In the \ref{sec:appendix}, we explore how changing the ranges for neutral, positive and negative papers affects our results that relate to a coarse level. In particular, we there define neutral papers as ones that fall into $(-0.2,+0.2)$, with corresponding implications for negative and positive papers. Overall, we find that our results are very similar for such slightly different range definitions.} We show the more detailed distribution of papers in terms of stance in \ref{fig:histogram-stance}.

\begin{figure*}[!htb]
  \centering
  \includegraphics[width=0.95\linewidth]{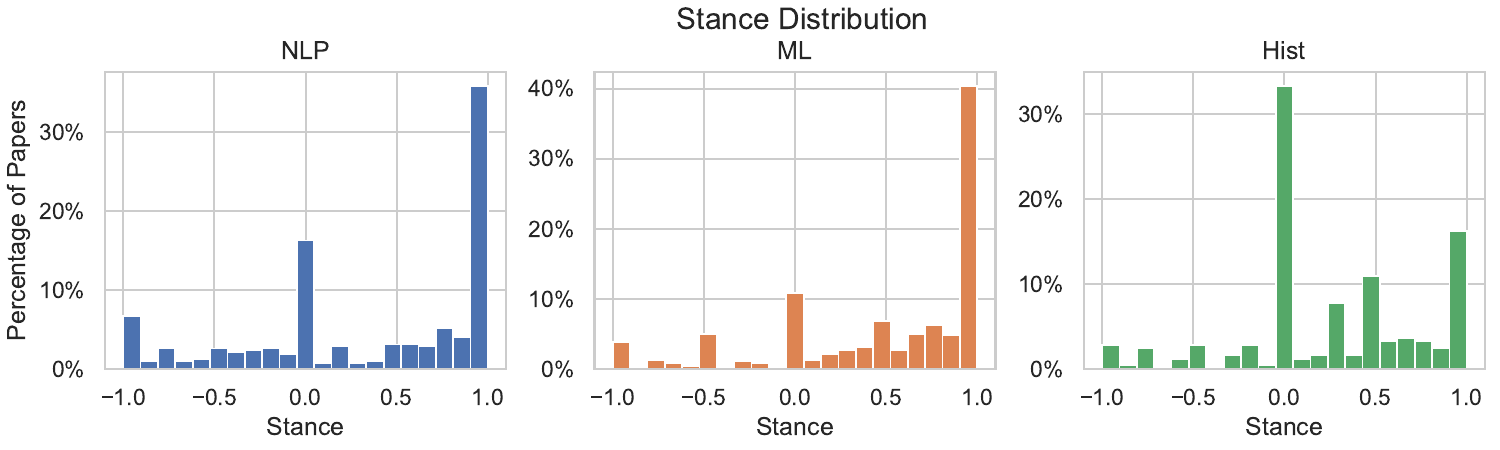}
  \caption{Stance distribution of the human-annotated dataset (Hist: historic papers).}
  \figurelabel{fig:histogram-stance}
\end{figure*}

This statistic does not reflect the true distribution of stance in our data (which is dominated by positive papers, as we will show below), as we oversampled negative papers using heuristics (e.g. looking for particular keywords in abstracts and titles such as \emph{fail} and \emph{limitation} and titles with question marks). Candidates for positive papers were randomly drawn. We did this in order to ensure  classifiers trained on datasets that are not too class imbalanced. Manipulating the (training) data distribution is a common approach to ensure better classifiers in the face of small minority classes \citep{9078901}; our heuristics-based approach also makes annotation much more efficient, as we would otherwise have to annotate considerably more data to obtain negative instances. We note that we \emph{evaluate} our models also on the `natural' data distribution, not only the skewed one; see \ref{sec:experiments}.

\subsection{Inter-annotator agreement}
We had up to 4 annotators annotate abstracts for stances. The annotators were computer science undergraduate students and one computer science faculty member from NLP. 71\% of the human-annotated data set is annotated by one, 24\% by two, 2\% by three, and 1\% by four annotators. This distribution reflects the fact that annotating all instances by all annotators would have been too costly and, given good agreements, also not necessary. Newly incoming annotators first annotated already annotated instances --- in order to measure agreement, i.e. their task understanding --- then proceeded to annotate independently. We label each abstract's stance as the average over all the annotators.

We measure agreement on stance annotation using Pearson's Correlation Coefficient, Cohen's Kappa Coefficient, and Krippendorff's Alpha Coefficient. The resulting Pearson correlations among all pairs of annotators (on a common set of instances) range from 0.64 to 0.94 (avg.: 0.77). On a coarse level with three stances, the kappa agreement is between 0.53 to 0.87 (avg.: 0.66). The alpha agreement among all annotators is 0.74 on a ratio scale. Overall, we thus have good agreement. We illustrate inter-annotator agreement (kappa, Pearson) and the number of common instances in \ref{fig:inter_annotator_agreement}.

\begin{figure}[!htb]
  \centering
  \includegraphics[width=0.72\linewidth]{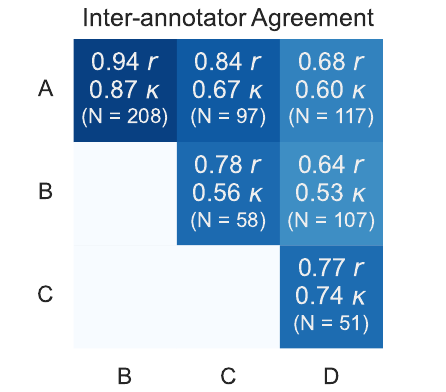}
  \caption{Inter-annotator agreement measured using Pearson's correlation coefficient (\textit{r}) and Cohen's kappa coefficient (\textit{$\kappa$}), and number of common instances (\textit{N}).}
  \figurelabel{fig:inter_annotator_agreement}
\end{figure}

\subsection{Historic versus modern data}
We refer to \NLP{} and \ML{} papers published before the year 2000 as \emph{historic} papers (\Hist{}) and papers published since 2000 as \emph{modern} papers. The historic dataset consists of 246 papers, of which 194 belong to \NLP{} and 52 to \ML{}. Modern \NLP{} consists of 578 and modern \ML{} of 726 papers.

\section{Model}\sectionlabel{sec:implementation}

For all our experiments, we use SciBERT \citep{SciBERT}. We feed each paper as concatenation of title and abstract separated by special tokens to SciBERT: \texttt{[CLS] <title> [SEP] <abstract>}. We set the maximum token length to 300, which is sufficient for most papers and a good compromise for efficient memory usage. We add a fully connected layer with one output neuron and linear activation on top of the pooled output to obtain a single prediction for the stance of a paper. Since we define stance as a value between $-$1 and +1, we clip the prediction to the desired range.

The model is fine-tuned with the following hyperparameters: a batch size of 8 or 16, a \emph{slanted triangular learning rate} \citep{ULMFiT} with a maximum learning rate of $1\times 10^{-5}$, $2\times 10^{-5}$, or $5\times 10^{-5}$, a warm-up ratio of 0.06, and linear decay. We train for 2, 3 or 4 epochs and optimize using Adam \citep{Adam} with an $\epsilon$ of $1\times 10^{-6}$, $\beta_1$ of 0.9, $\beta_2$ of 0.999, and the mean squared error (MSE) as the loss function.

For our experiments in \ref{sec:experiments}, we train 18 models by performing a full grid search over the specified hyperparameters and keep the best model based on the MSE on the dev set. We repeat this five times and calculate performance metrics (cf. \ref{sec:setup}) as the average score of those models.

\section{Experiments}\sectionlabel{sec:experiments}

In the following, we first verify the reliability of our stance detection model described in \ref{sec:implementation}. To do so, we assess its \emph{cross-domain} and \emph{in-domain performance} and compare it with several baselines. Once the quality of the model is assured, we apply it large-scale to determine trends over time and venues in \ref{sec:analysis}.

\subsection{Experimental set-up}\subsectionlabel{sec:setup}
\subsubsection{Metrics}
We use various metrics to evaluate our models. The \textbf{coefficient of determination (R²)} is similar to the MSE but also takes the distribution of the data into account, which makes it more informative and truthful than the MSE \citep{R2IsBetterThanMSE}. A model that always predicts the expected value has an $R^2$ score of 0. The range of the metric is $(-\infty, 1]$. The \textbf{macro F1 score} is a standard metric to assess the quality of multi-class classification which can take class-imbalance into account. We compute the macro F1 score on coarse-grained stance labels (positive, negative, neutral), see above. We also calculate the \textbf{F1 score} for the labels individually. The \textbf{natural macro F1 score} samples papers according to the natural distribution of the data (i.e. mostly positive), as predicted by our best performing model. To do this, we draw the test sets randomly (from the existing test sets) according to the true distribution\footnote{We are aware that the natural distribution, as we define it, relies on our models' predictions. As these are averages over many instances, we think that our models' predictions can be relied upon at aggregate level; see our results below.} and then calculate the macro F1 score on the three labels. Since randomness is involved, we repeat this 1 k times and average the scores.

\subsubsection{Baselines}
We compare our models with simple baselines. 
{\bf POS}: always predict a positive stance +1;
{\bf ZERO}: always predict a neutral stance 0;
{\bf NEG}: always predict a negative stance $-$1; 
{\bf AVG}: always predict the average of manual annotations.

\subsubsection{Cross-domain experiments}
Due to the exponential growth of science, our model is mostly trained on more recent data. However, we also want to make sure that we obtain reliable predictions, e.g. for past papers. As a consequence, we first evaluate our model in a cross-domain setting. In this, we train our model on papers from different time stamps or domains and evaluate on a respective out-of-domain test set. For the \textbf{source data}, we set the train--dev split ratio to 0.7 and 0.3, and we use the whole annotated data for the \textbf{target data}.

\subsubsection{In-domain experiments}
We also perform in-domain tests where we train and test on the same domain of data (e.g.\ modern \NLP{}). We set the train--dev--test ratio to 0.6/0.1/0.3.

\subsubsection{Combined experiments}
We create combined train and dev sets which consist of \NLP{}, \ML{} and \Hist{} papers and evaluate on each domain individually. We set the train--dev--test ratio to 0.6/0.1/0.3.

\subsubsection{Data}
As mentioned in \ref{sec:procedure}, we divided our human-annotated dataset into three groups. The human-annotated \NLP{} dataset consists of a total of 578 papers (60\% positive, 16\% neutral, 24\% negative). The \ML{} dataset consists of 726 annotated abstracts (75\% positive, 11\% neutral, 14\% negative). The \NLP{} portion of the \Hist{} dataset has 194 annotated samples (58\% positive, 30\% neutral, 12\% negative); the \ML{} portion of the \Hist{} dataset contains 52 papers (31\% positive, 46\% neutral, 23\% negative).

\begin{figure*}[!htb]
    \centering
    \includegraphics[width=1.0\linewidth]{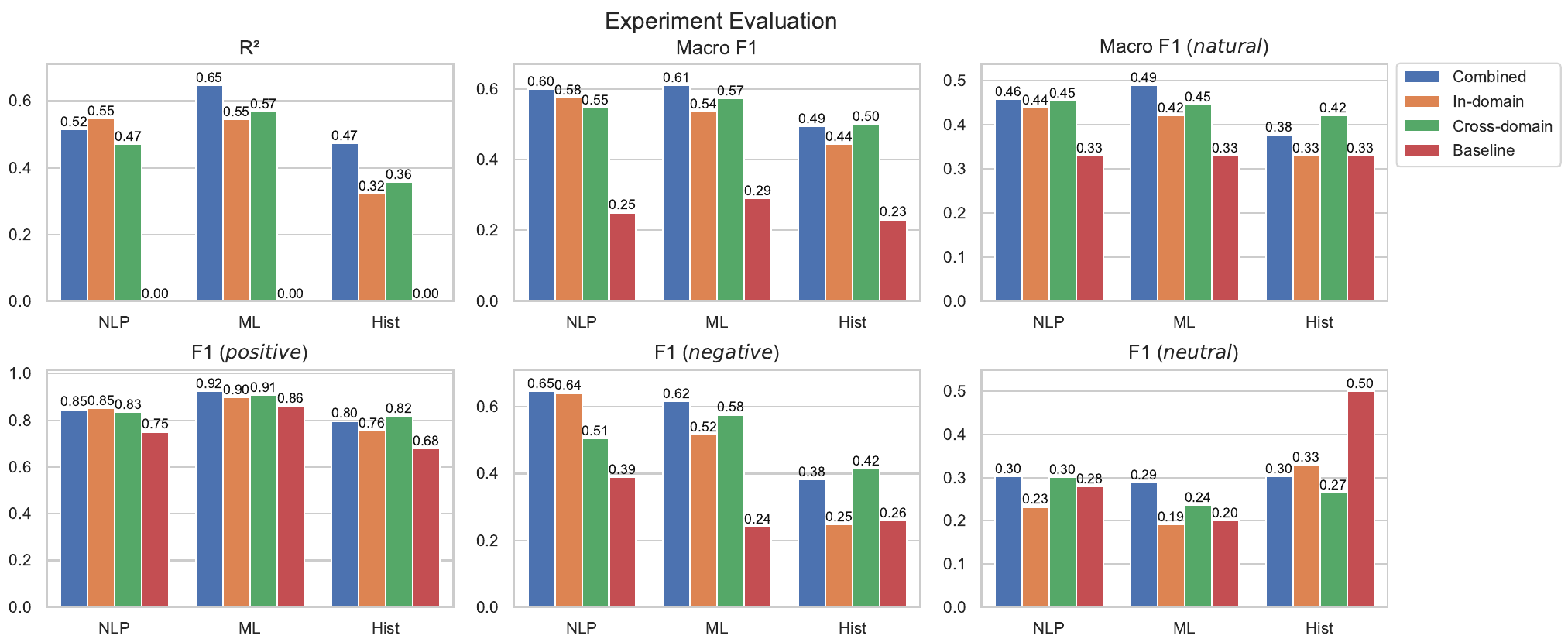}
    \caption{Different metric scores for combined, in-domain and cross-domain evaluation. The x-axis represents the test data. and the coloured bars represent different train sets. Cross-domain is the train set which consists of the two remaining datasets. The individual F1 scores in the bottom half refer to the macro F1 score in the top middle. We choose the best baselines for comparison; AVG for R², macro F1, macro F1 (\textit{natural}), and F1 (\textit{positive}); ZERO for F1 (\textit{neutral}); and NEG for F1 (\textit{negative}).}
    \figurelabel{fig:experiment_evaluation}
\end{figure*}

\subsection{Results}\subsectionlabel{sec:results}
Results are shown in \ref{fig:experiment_evaluation}. We observe clear trends: in-domain and cross-domain performance are typically close, but cross-domain performance is better on average (note that cross-domain uses between 1.3 and 6.2 times as much training data as in-domain; cf.\ \ref{tab:experiment_data_sizes}). The model trained on combined data, which uses even more data, outperforms in-domain and cross-domain results. In-domain and cross-domain performance on \Hist{} is lowest, which is not surprising as this dataset is smallest in size and assumedly has largest divergence to the modern datasets, due to temporal divergence \citep{lazaridou2021pitfalls}. On average, the models are best in predicting the positive class, and worst in predicting the neutral class. It is interesting to note that a model trained on a combination of all sources and time periods performs best. It even leads to good results on the \Hist{} portion of our data, which is why we use it for the analysis below.

\begin{table*}[!htb]
    \centering
    \begin{tabular}{l|rrr|rrr|rrr}
        \toprule
        \multirow{2}{*}{} & \multicolumn{3}{c|}{\textbf{Combined}} & \multicolumn{3}{c|}{\textbf{In-domain}} & \multicolumn{3}{c}{\textbf{Cross-domain}} \\
         & \textit{train} & \textit{dev} & \textit{test} & \textit{train} & \textit{dev} & \textit{test} & \textit{train} & \textit{dev} & \textit{test} \\
        \midrule
        \NLP{}  & 931 & 156 & 173  &  347 &  58 & 173  &  680 & 292 & 578 \\
        \ML{}   & 931 & 156 & 217  &  436 &  73 & 217  &  577 & 247 & 726 \\
        \Hist{} & 931 & 156 &  73  &  148 &  25 &  73  &  913 & 391 & 246 \\
        \bottomrule
    \end{tabular}
    \caption{Sizes of the \textit{train}, \textit{dev}, and \textit{test} sets of the experiments in \ref{sec:experiments}, whose results are shown in \ref{fig:experiment_evaluation}, for combined, in-domain and cross-domain evaluation and \NLP{}, \ML{}, and \Hist{} test data.}
    \tablelabel{tab:experiment_data_sizes}
\end{table*}

\subsection{Error analysis}\subsectionlabel{sec:error-analysis}
We further assess the quality of our best performing combined model by comparing the human annotated and predicted stance in the test set using the confusion matrix shown in \ref{fig:confusion_matrix}. Positive papers are correctly predicted in most cases. Neutral and negative papers are more frequently confused as positive and when the model predicts a paper to be negative the true class is sometimes neutral.

\begin{figure}[!htb]
    \centering
    \includegraphics[width=0.72\linewidth]{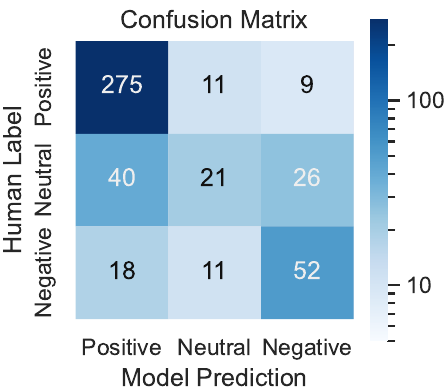}
    \caption{Comparison of human-annotated and model-predicted stance of our best model from the experiments in \ref{sec:experiments}.}
    \figurelabel{fig:confusion_matrix}
\end{figure}

In \ref{sec:appendix} \ref{tab:papers}, we illustrate sample predictions of our model, randomly sampled from a predicted stance of more than 0.8 (very positive papers) or below $-$0.7 (very negative papers). We note that the predicted values look very plausible overall. However, especially for the positive papers, the model misses some negative elements, thus overestimates their positivity; arguably the negative papers are also slightly more negative than the model predictions.

We further conduct a manual analysis on papers that have a large discrepancy between predicted and human stance. In these cases, we find that the model often predicts very negative papers as very positive. This may be related to positive papers being the majority class. In several cases, the model was also seemingly led astray by small positive contributions, especially in last sentences, and by superficial cues indicating positive contributions such as \emph{we propose}. One pattern for the misclassified papers is strong criticism followed by the release of a new dataset indicated in the last sentence. In some of the error cases, the gold standard is also incorrect or too extreme, e.g. too strongly negative.

\subsection{Relation between stance and sentiment}\subsectionlabel{sec:stance-vs-sentiment}
We compare the stance predictions of our model with the predictions of several sentiment analysis tools to assess how similar our concept of `stance' and the concept of `sentiment' are. We test (i) a RoBERTa \citep{RoBERTa} model for sentiment analysis, called SiEBERT \citep{HARTMANN2022}, trained on 15 sentiment data sets from diverse text sources (Tweets, reviews, etc.), (ii) VADER \citep{VADER}, a rule-based model for sentiment analysis of social media text, (iii) TextBlob\footnote{\url{https://textblob.readthedocs.io/}}, a text processing library, and (iv) a valence/arousal lexicon \citep{Warriner2013-vw}. We evaluate the similarity of stance and sentiment using Pearson's correlation coefficient, which ranges from 0.01 to 0.27. We interpret the low correlation in two ways: (i) it may be the result of different domains for the training data, which for many sentiment models are texts from social media, while we use scientific texts; (ii) it indicates that our definition of `stance' involves different nuances than simple `sentiment'. For example, text such as `we propose a new model' indicates positivity in our context, but may be considered neutral sentiment. Similarly, `we identify limitations in existing works' may be considered neutral sentiment polarity, but indicates negativity in our context.

\begin{figure*}[!htb]
    \centering
    \subfloat[Overall distribution of stance values.]{
        \includegraphics[width=0.5\linewidth]{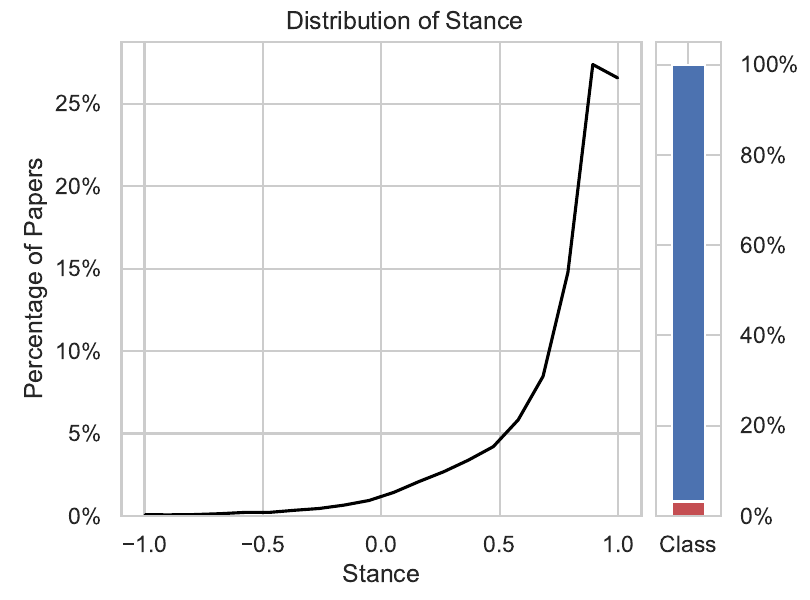}
        \subfigurelabel{fig:distribution_of_stance}
    }
    \subfloat[Distribution of stance values for each domain.]{
        \includegraphics[width=0.5\linewidth]{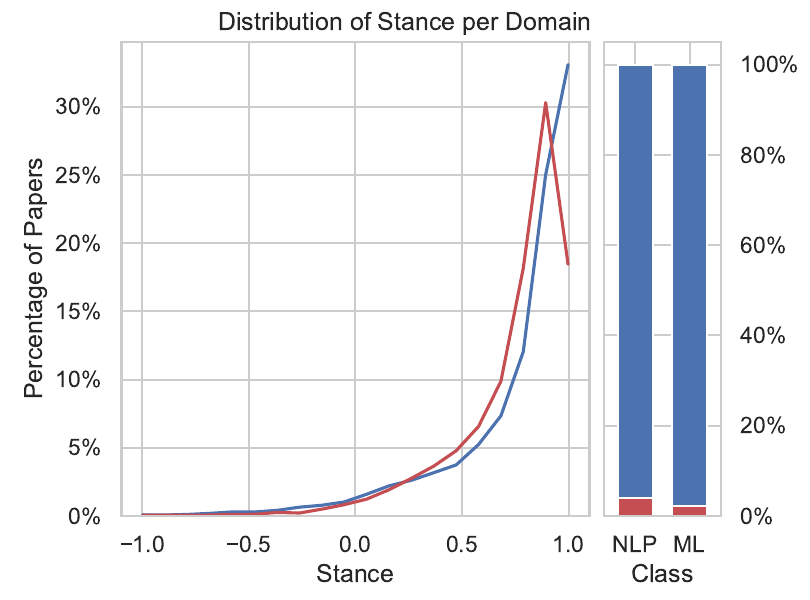}
        \subfigurelabel{fig:distribution_of_stance_per_domain}
    }
    \caption{Distribution of stance values. The line plot on the left shows how the stance values are distributed; the bar plot on the right shows the percentage of papers with positive and negative stance.}
\end{figure*}

\section{Analysis}\sectionlabel{sec:analysis}

We analyse large-scale trends from the combined model's predictions and smooth the graphs with Gaussian blurs.\footnote{The observed trends correlate between our model and a model trained as a reference on the \emph{full dataset} with 1.5 k papers (no test split) with a Pearson's Correlation Coefficient from 0.76 to 0.99 (avg.: 0.96)}. We further use Welch's \ttest{} \citep{WelchTest} and the Kruskal--Wallis \Htest{} \citep{KruskalWallisTest} to detect differences in the distributions and to test our hypotheses; we report the achieved significance levels in parentheses.

Our first questions connect to the recent paper of \citet{bowman-2022-dangers}, who observes `a wave of surprising negative results' in recent years in the NLP community, partly confirming his evidence from selected case studies with our large-scale predictions (but partly also putting it in perspective).

\subsection{`Are there more positive or more negative papers?'}
The histogram of stance values predicted by the model, aggregated over all venues and years, is visualized in \ref{fig:distribution_of_stance}. It shows that most papers have a positive stance and that the more negative the stance gets, the fewer papers there are. Less than 4\% of all papers have a negative stance and more than ¾ of all papers have a stance of greater than or equal to 0.6.

\subsection{`Are \NLP{} papers more positive/negative than \ML{} papers?'}
Both hypothesis tests, the \ttest{} and the \Htest{}, show with a significance level of 0.01\% that the distribution of predicted stance values differs between \NLP{} and \ML{}. \ref{fig:distribution_of_stance_per_domain} shows the histogram of the predicted stance values, aggregated over all years, for both datasets. The distribution is similar to the overall trend, but \ML{} has more papers with stance values between 0.5 and 0.8, whereas \NLP{} has more papers with a stance of 1.0. Overall, 3.9\% of all \NLP{} papers and 2.3\% of all \ML{} papers exhibit a negative stance, which makes \NLP{} more negative than \ML{}.

\subsection{`Did AI get more positive/negative recently?'}
We analyse the development of the \emph{average stance value} over time in \ref{fig:average_stance_per_year_and_domain}. This shows that the average stance is always positive with a minimum value of 0.34 for \NLP{} in the 1980s. When our \ML{} dataset started in 1989 it was more positive than \NLP{} (p-value 0.01\%, \ttest{}). In the late 1990s, the stance of papers in \NLP{} and \ML{} came closer together, and in the 2000s (when NAACL and SemEval were first held) \NLP{} took over and became slightly more positive than \ML{} (p-value 0.01\%, \ttest{}). The positiveness reached its peak around 2015 with an average stance above 0.80 for \NLP{}. It then started to get more negative for \NLP{} which means that the field of \NLP{} got less positive recently. The \ML{} domain, however, got more positive with a maximum stance of 0.81 in 2021.

Overall, the average stance of both \ML{} and \NLP{} increased substantially from the mid-1980s to the 2020s. This means that papers got more positive on average, i.e. build upon another and report better and better results (i.e. new state-of-the-art performances), confirming the observation that \ML{} and \NLP{} have become `rapid discovery sciences' \citep{MeasuringTheEvolution}.

\begin{figure}[!htb]
    \centering
    \includegraphics[width=\linewidth]{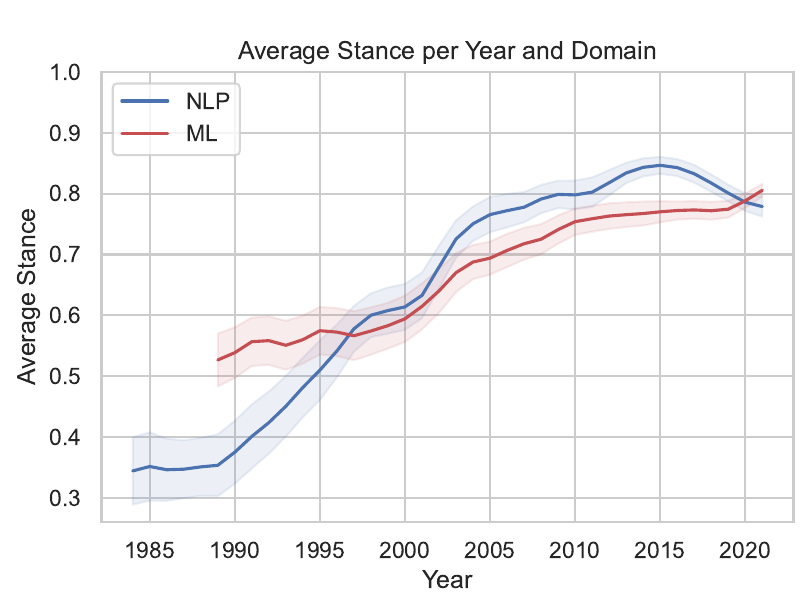}
    \caption{Development of the average stance value over the years for both domains, average stance and 95\% confidence interval.}
    \figurelabel{fig:average_stance_per_year_and_domain}
\end{figure}

We further analyse whether the increase in positiveness from 1990 to 2010 and the decrease in positiveness in the most recent years for \NLP{} comes from more positive/negative papers overall or from positive/negative papers getting less/more positive/negative. \ref{fig:negative_stance_per_year_and_domain} shows how many papers have a negative stance in each year. We observe that negativity has peaked in the 1980s and 1990s for \NLP{} and \ML{}, respectively. There was then a continuous downward trend in negativity until the 2010s. In the recent years, negativity has increased for both domains, but considerably more sharply so for \NLP{}, from 2\% of all papers in 2015 to 4\%.

\begin{figure}[!htb]
    \centering
    \includegraphics[width=\linewidth]{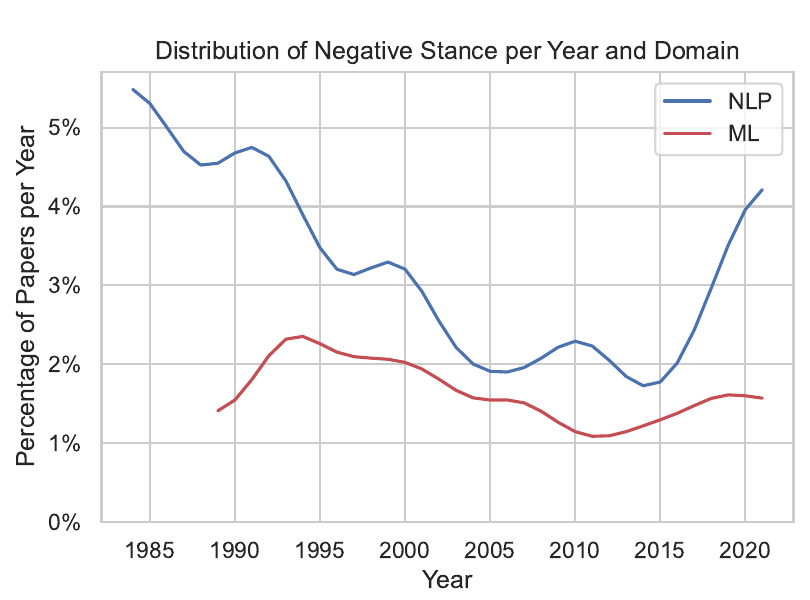}
    \caption{Percentage of papers with a negative stance, calculated as the number of negative papers divided by the total number of papers in each year, for both domains.}
    \figurelabel{fig:negative_stance_per_year_and_domain}
\end{figure}

\ref{fig:average_positive_and_negative_stance_per_year_and_domain} shows the \emph{average stance value of all positive papers} and the \emph{average stance value of all negative papers} over time. The development of the average positive stance value is very similar to the development of the average stance value overall. By contrast, the average negative stance value has a decreasing trend for both datasets which means that negative papers have become more negative over time.

\begin{figure*}[!htb]
    \centering
    \includegraphics[width=0.95\linewidth]{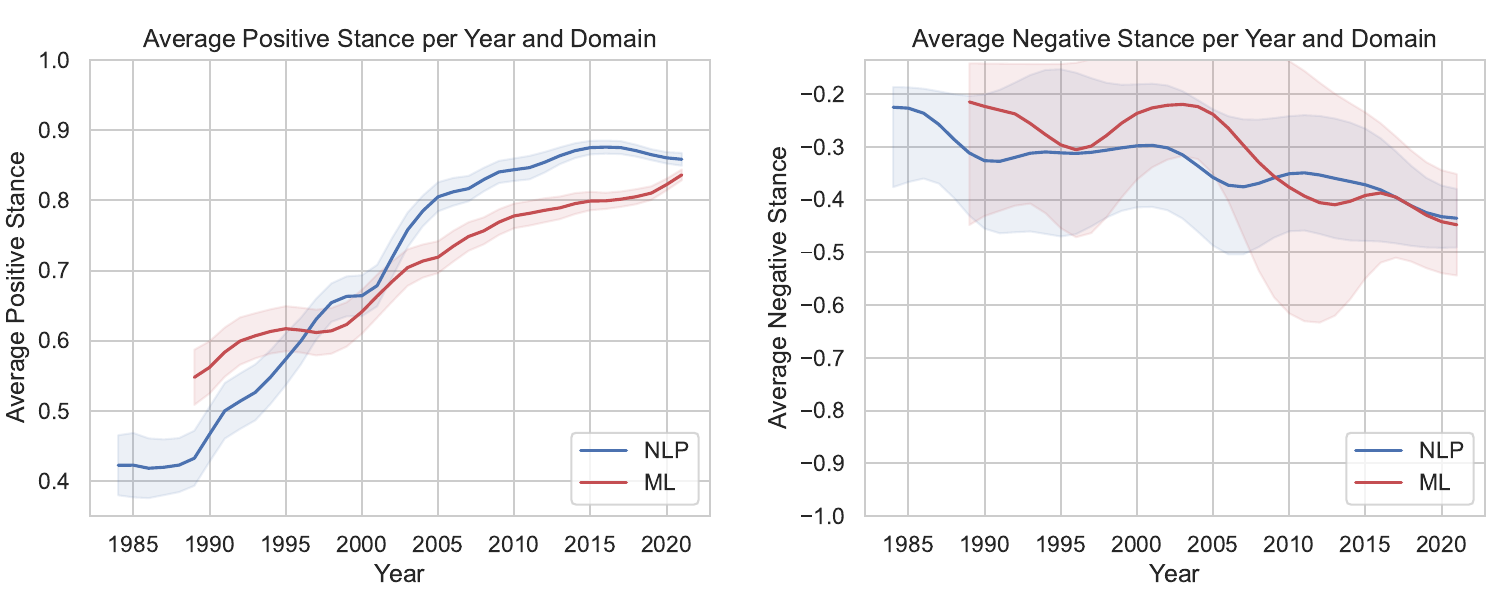}
    \caption{Distribution of the average stance value of papers with a positive (left) and a negative (right) stance over time for both domains, average stance and 95\% confidence interval. The values indicate the average positiveness/negativeness of positive/negative papers.}
    \figurelabel{fig:average_positive_and_negative_stance_per_year_and_domain}
\end{figure*}

Finally, we analyse trend curves for individual venues, visualized in \ref{fig:negative_stance_per_year_and_venue}. The trend towards more negative papers in the most recent years is visible for most venues, especially ACL, EMNLP, COLING, NAACL, CoNLL and ICML. TACL has the sharpest increase. SemEval and AAAI do not follow this trend, however. Many venues were more negative before the 2000s and least negative in the 2000s. The CL journal is a noticeable outlier: it is the most negative venue with up to 16\% negative papers (p-value 0.01\%, \ttest{}); we note that it is also the only journal in our dataset besides TACL.

\begin{figure}[!htb]
    \centering
    \includegraphics[width=\linewidth]{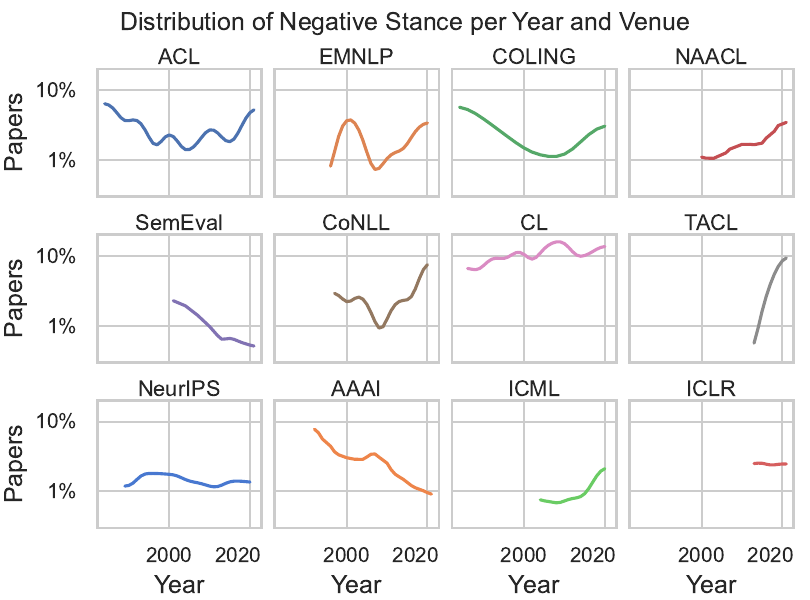}
    \caption{Percentage of negative papers, calculated as the number of negative papers divided by the total number of papers in each year, for each venue on a logarithmic scale.}
    \figurelabel{fig:negative_stance_per_year_and_venue}
\end{figure}

\subsection{`Do positive/negative papers receive more/fewer citations?'}
\ref{fig:normalized_number_of_citations_per_stance} shows how many citations a paper with a certain stance value has received in comparison with papers published in the same year. We compare citation counts using normalized values that indicate how many citations more or less a paper has received in comparison with the average number of citations a paper published in the same year has received, measured in multiples of the standard deviation of citation counts in that year. Positive values indicate more citations than the average, negative values indicate fewer citations. The graph shows that papers with a negative stance of $-$0.3 or less receive more citations than the average paper in the same year (p-value 3\%, \ttest{}); very negative papers receive even more citations. By contrast, a paper with a positive stance receives less citations on average (p-value 6\%, \ttest{}), but very positive papers with a stance value of more than 0.8 receive slightly more citations (p-value 3\%, \ttest{}).

\begin{figure}[!htb]
    \centering
    \includegraphics[width=\linewidth]{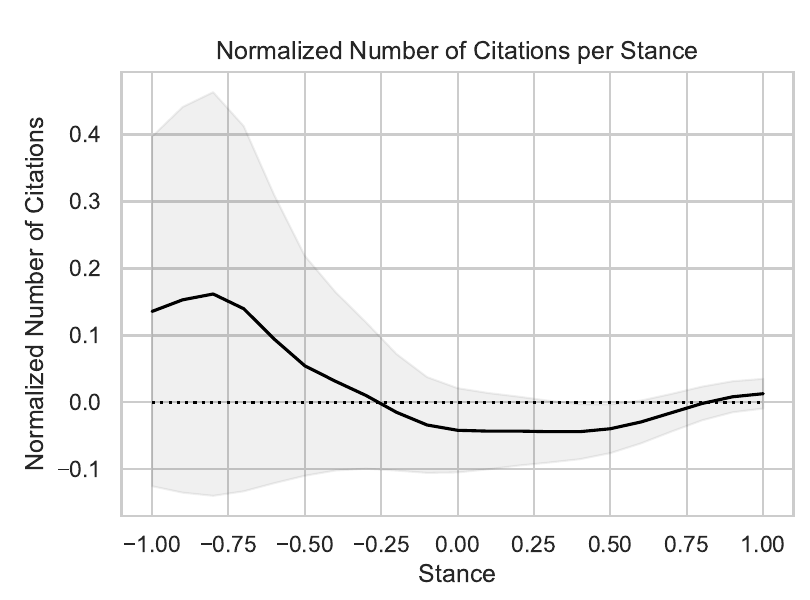}
    \caption{Normalized number of citations a paper with a certain stance value has received, for both domains combined, average number of normalized citations and 95\% confidence interval. Normalized values indicate how many citations more or less a paper has received in comparison with the average number of citations a paper published in the same year has received, measured in multiples of the standard deviation of citation counts in that year.}
    \figurelabel{fig:normalized_number_of_citations_per_stance}
\end{figure}

Similar results can be found by analysing the individual domains, \NLP{} and \ML{}, separately, which is shown in \ref{fig:normalized_number_of_citations_per_stance_and_domain}. The domain of \ML{} is more extreme than \NLP{} in that negative papers with a stance of $-$0.5 or less receive even more citations (p-value 6\%, \ttest{}) and positive papers with stance values from 0.1 to 0.7 even fewer (p-value 0.1\%, \ttest{}).

\begin{figure}[!htb]
    \centering
    \includegraphics[width=\linewidth]{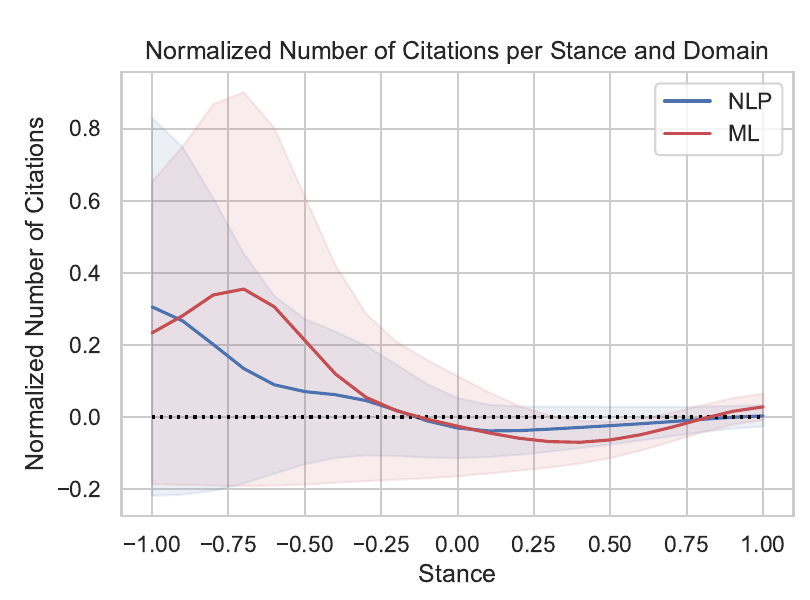}
    \caption{Normalized number of citations a paper with a certain stance and domain has received, average number of normalized citations and 95\% confidence interval.}
    \figurelabel{fig:normalized_number_of_citations_per_stance_and_domain}
\end{figure}

Overall, this shows that papers of negative stance seem to attract more citations than papers of neutral or positive stance, indicating that they receive more attention and have larger effect on the community. Together with \citet{catalini:2015}, this means that a paper of negative stance receives more citations but decreases the citation counts of papers that it cites negatively, which may indicate that it shifts attention from those papers to itself. The fact that papers of negative stance receive more citations would make stance also suitable for inclusion as feature in models that predict citation count \citep{yan2011citation}.

However, we acknowledge that also other factors may be at work here, e.g. that stance may only be a confounding variable. For example, it could be that high-prestige authors (e.g. measured in terms of their h-indices) tend to be over-represented in papers of negative stance; high-quality authors, in turn, may attract more citations (potentially because they write better papers), which could explain the link between citations and stance. To analyse the relationship between stance and citations in more depth, we performed a linear regression similar to \citet{sienkiewicz2016impact} by measuring various factors (see (a)--(d) below), including the length and complexity of titles and abstracts, as well as a paper's stance, to predict its citation count. We could not reproduce their results because the goodness-of-fit of the regression was low, indicating that linear models are not the adequate choice in our case. Instead, we analyse which factors differ between very positive (greater than or equal to $0.8$), very negative (less than or equal to $-0.8$), and neutral papers ($\in (-0.1, +0.1)$) on a subset of 39 k papers for which we have all required metadata, which leads to 24.1 k very positive, 69 very negative, and 874 neutral papers. Similarly to \citet{sienkiewicz2016impact}, we explore the following factors: (a) \textbf{length} (characters in title, and words in abstract), (b) \textbf{complexity} (Herdan's C index \citep{Herdan1960LanguageAC}, z-score \citep{Gerlach2014ScalingLA}, and Gunning fog index \citep{Gunning1952TheTO}), (c) \textbf{sentiment} (average valence/arousal \citep{Warriner2013-vw}), and (d) \textbf{author information} (number of authors, mean/minimum/maximum h-index of authors\footnote{We use h-indices of a paper's authors we collected from Semantic~Scholar in late 2021, i.e. these are not historic h-indices from the time of publication of a paper.}) for papers of positive, negative, and neutral stance. We standardize all factors per venue and show the mean of each factor together with the 95\% confidence interval in \ref{fig:confidence-interval-analysis}. While the length of the title does not reveal a clear trend, longer abstracts are more common in neutral papers. The variance for complexity factors is generally too high to draw conclusions, but titles with a high z-score, i.e. high complexity, tend to be more common in very negative papers. Very negative papers have higher arousal and neutral papers have lowest arousal. Author information separates positive, negative and neutral papers best. In contrast to our initial hypothesis expressed above, very negative papers have authors with lower (current) h-indices.

\begin{figure*}[!htb]
    \centering
    \includegraphics[width=\linewidth]{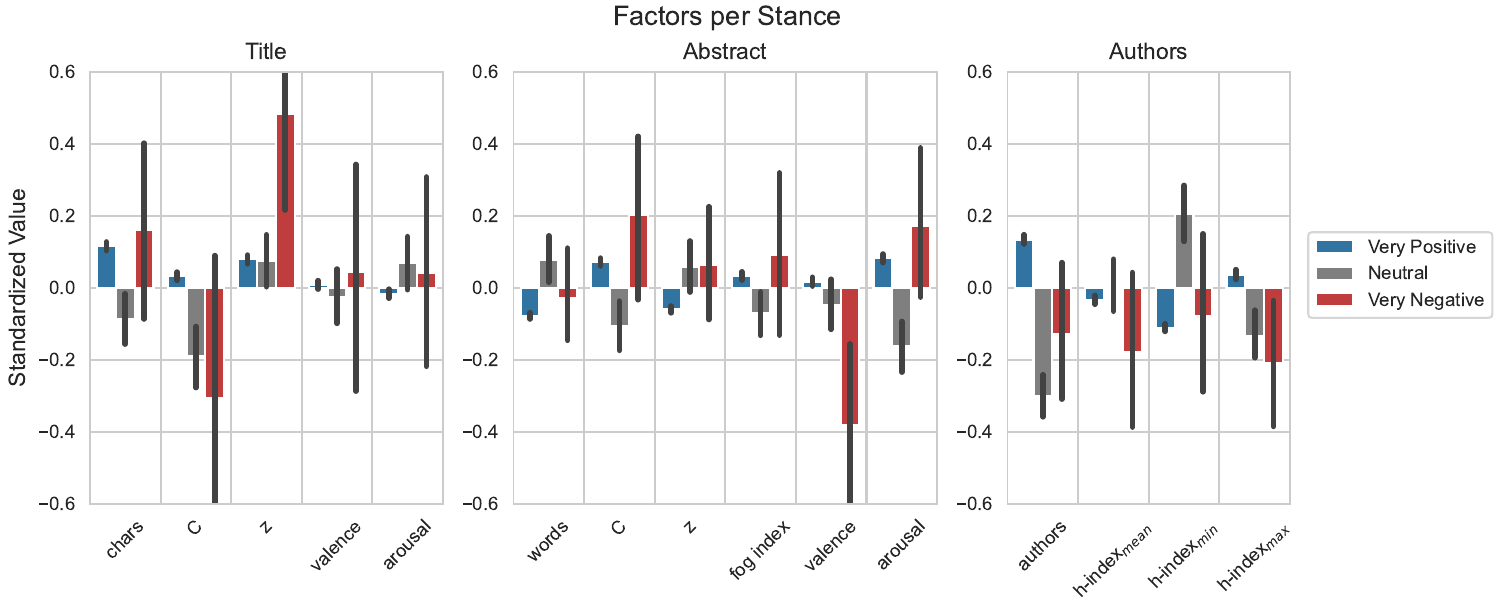}
    \caption{Standardized values for various metrics of very positive, very negative, and neutral papers, average value and 95\% confidence interval. Metrics are explained in the main text.}
    \figurelabel{fig:confidence-interval-analysis}
\end{figure*}

\subsection{`Do positive/negative papers have lower/higher acceptance chances?'}
We use 5,453 accepted and 14,974 rejected papers from the ICLR conferences in 2013 and 2017--2021, collected from OpenReview\footnote{\url{https://openreview.net/}}, and from ACL, EMNLP, NAACL, NeurIPS, AAAI, ICML, and ICLR over the years 2007--2017 as collected by \citet{PeerReviewsDataset}. The \ttest{} and the \Htest{} show with a significance level of 0.01\% that the distribution of predicted stance values differs between accepted and rejected papers. \ref{fig:acceptance_rate_per_stance} shows how many papers with a certain stance value were accepted. The trend indicates that papers with a negative stance of $-$0.6 or less have higher acceptance rates (p-value 1\%, \ttest{}). The acceptance rate for papers with stance values between $-$0.6 and 0.8 is lower than the overall acceptance rate (p-value 0.01\%, \ttest{}), but for very positive papers, the acceptance rate is slightly higher again (p-value 0.01\%, \ttest{}).

\begin{figure}[!htb]
    \centering
    \includegraphics[width=\linewidth]{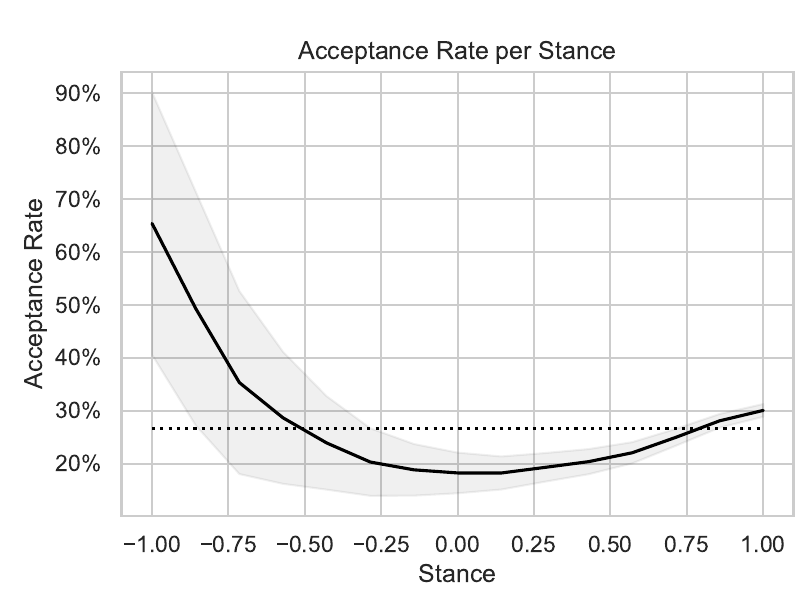}
    \caption{Acceptance rate of a submitted paper with a certain stance value, average acceptance rate and 95\% confidence interval. The dotted line indicates the overall acceptance rate of 26.7\%.}
    \figurelabel{fig:acceptance_rate_per_stance}
\end{figure}

We also calculate the acceptance rates for two separate time spans, 2007--2014 and 2015--2021, as shown in \ref{fig:normalized_acceptance_rate_per_stance_and_year}. The \ttest{} and the \Htest{} show that the acceptance chances are different in the two time spans with a significance level of 0.01\%. For the most recent years 2015--2021, the trend is similar to the overall trend, including that papers with a very positive stance of more than 0.8 are more likely to be accepted (p-value 0.1\%, \ttest{}), but those papers do not have much higher chances. However, the acceptance rates in earlier years 2007--2014 were different. Papers with a very positive stance of 0.8 or more used to have better acceptance chances than very positive papers in 2015--2021 (p-value 0.1\%, \ttest{}). This is consistent with the trend over time, which shows that fewer papers were negative in the years 2007--2014 than in 2015--2021, implying a bias: positive papers were more popular back then and therefore more positive papers were more likely to be accepted.

\begin{figure}[!htb]
    \centering
    \includegraphics[width=\linewidth]{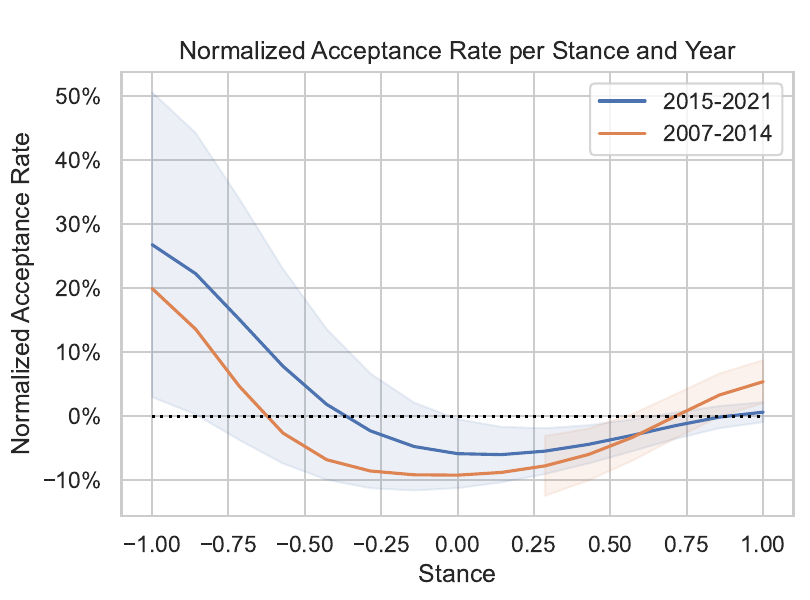}
    \caption{Normalized acceptance rate of a submitted paper with a certain stance value for two time spans, average acceptance rate and 95\% confidence interval. Normalized values indicate how many percentage points more or less a paper with a certain stance value is likely to be accepted in comparison with the average acceptance rate in each time span. The dotted line indicates the average acceptance rate.}
    \figurelabel{fig:normalized_acceptance_rate_per_stance_and_year}
\end{figure}

\section{Concluding remarks}\sectionlabel{sec:conclusion}

We analysed stance in abstracts of scientific publications, where authors position themselves positively or negatively (with respect to related work). We annotated over 1.5 k abstracts from ML and NLP venues and trained a SciBERT model on a subset of the annotated abstracts, verifying that the model is of sufficiently high quality for the task. We then used this model to automatically predict the stance of a paper based on its title and abstract. We applied the model large-scale to a collection of 41 k scientific publications in the domain of NLP and ML from the years 1984 to 2021 to enable large-scale analysis.

The analysis revealed that the majority of papers in the past and today have a positive stance, that the average stance has substantially increased over time, yielding support for the hypothesis that ML and NLP have become `rapid discovery sciences', and that the ML domain is more positive than the NLP domain. Scientific publications used to have a more negative stance in the early days, then became very positive until they started to get more negative again recently. Overall, publications also got more extreme over time, which means that positive papers became more positive and papers with a negative stance more negative. We found (very) negative papers to be more influential than (mildly) positive ones in terms of citations they receive and more likely to be accepted to NLP/ML venues.

We believe that NLP/ML turned more positive when the fields became more statistically solid, starting from the 1990s, and when authors started to build on the existing literature, with a peak of positivity in the mid-2010s, the beginning of the deep learning revolution. This hype has apparently also led to a recent increase in negativity, when papers started to challenge the validity of some of the claims \citep{Marcus2018DeepLA}, when issues of adversarial robustness \citep{kurakin2018adversarial} and reproducibility \citep{post-2018-call,Chen2022ReproducibilityIF} became apparent  and people began to question evaluation frameworks \citep{Reimers2017ReportingSD}.

Our results also inform the recent work of \citet{bowman-2022-dangers}, who warns of the dangers of (what he calls) `underclaiming papers' (which are negative papers in our terminology) by providing quantitative measures of negative papers over time. Given that negative papers tend to receive more attention (in terms of citations), we point out that they may also be key factors in improving the status-quo (cf.\ also \citep{catalini:2015}), which highlights their positive contribution to the scientific process. We also note that while negative papers have indeed sharply increased in numbers in the NLP domain (at least) recently, from a historical perspective, they are still on relatively low level.

Future work should address other scientific disciplines beyond NLP and ML for a broader scientific trend analysis, examine the correlation between overall stance of a paper and individual (negative) citations in its related work sections,\footnote{We conducted a small-scale experiment on 20 papers whether a negative paper actually addresses previous work in a negative way by comparing a paper's stance, predicted using title and abstract, with the stance of the citations, given in the related work or background section of a paper. To do this, we randomly selected 10 very positive and 10 very negative papers based on our model's predictions and manually annotated the citations (using citation context as in \citet{lauscher2021multicite}). In the sample, we find that a paper's stance and the difference between the amount of positive and negative citations are correlated with a Pearson's correlation coefficient of 0.62. This means that the stance of a paper may indeed be reflected in the way the authors frame  citations in their related work.} annotate word-level rationales for our sentence-level scores, assess the correlation between stance and socio-demographic factors (gender, nationality, affiliation, h-index etc.) and analyse how negative papers may potentially transform a field.

We release our data, code and model on GitHub.\footnote{\url{https://github.com/DominikBeese/DidAIGetMoreNegativeRecently}}

\section{Limitations}\sectionlabel{sec:limitations}

Limitations of our work especially relate to data quality of historic papers. We find that historic data is more prone to error because of (i) papers with unconventional form (e.g.\ missing abstracts, OCR errors) and (ii) the decreased quality Science~Parse on such papers. This leads to increased uncertainty especially for papers published before the year 2000.

To investigate this more thoroughly, we conducted a manual analysis of 800 papers where we annotated if the paper does really have an abstract and if the abstract was identified correctly by Science~Parse, i.e.\ if beginning and end of the abstract are correct. \ref{fig:data_quality_errors} shows that some historic papers do not have an abstract, while this is 
very rarely 
the case for data from 2010 onwards. Likewise, more recent papers are more often parsed correctly: overall, there are clear downward trends of wrongly parsed papers and papers without abstracts over time.

We have a data release v1.1 in which papers without abstracts have been removed and incorrectly parsed abstracts have been corrected (for \NLP{}, \ML{}, and the human-annotated dataset, i.e.\ train/dev/test splits).\footnote{\url{https://github.com/DominikBeese/DidAIGetMoreNegativeRecently/tree/data-update}}

\begin{figure}[!htb]
    \centering
    \includegraphics[width=\linewidth]{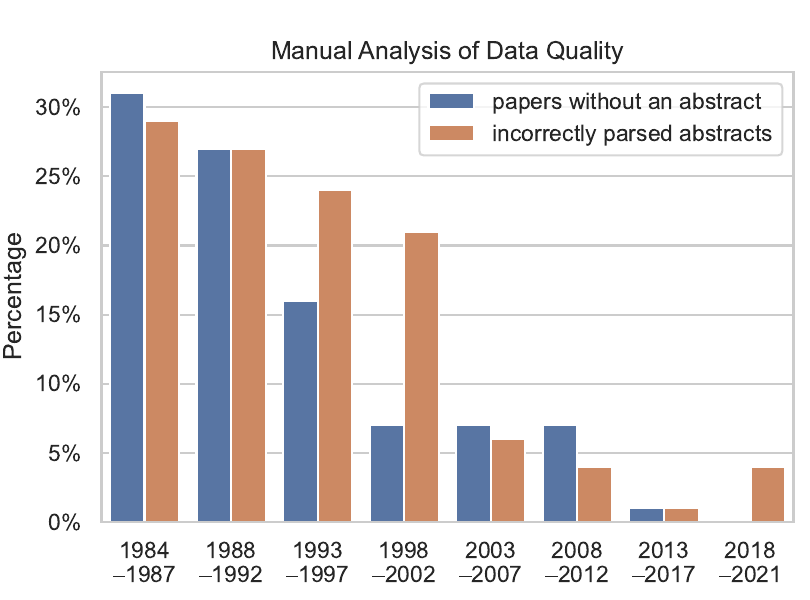}
    \caption{Percentage of erroneous data relating to (i) papers that do not have an abstract and (ii) papers where Science~Parse could not correctly identify the beginning and/or end of the (existing) abstract.}
    \figurelabel{fig:data_quality_errors}
\end{figure}

\vspace{15pt}
{\footnotesize

\noindent\textbf{Data accessibility.}
Data and relevant code for this research work are stored in GitHub: \url{https://github.com/DominikBeese/DidAIGetMoreNegativeRecently} and have been archived within the Zenodo repository: \url{https://doi.org/10.5281/zenodo.7590214} \citep{zenodo}.

\noindent\textbf{Authors' contributions.}
D.B.: data curation, formal analysis, methodology, software, visualization, writing--original draft, writing--review and editing;
B.A.: data curation, formal analysis, software, writing—original draft;
G.G.: data curation, formal analysis, software, writing—original draft;
S.E.: conceptualization, methodology, project administration, supervision, writing--review and editing.
All authors gave final approval for publication and agreed to be held accountable for the work performed therein.

\noindent\textbf{Conflict of interest declaration.}
We declare we have no competing interests.

\noindent\textbf{Funding.}
The last author was supported by DFG grant no. EG 375/5--1.

\noindent\textbf{Acknowledgements.}
We thank all the anonymous reviewers for their feedback and suggestions which greatly helped improve and clarify the paper. We also thank Farah Shahid for her initial help in annotating the data.

}

\bibliography{bibliography}

\begin{thebibliography}{74}
\expandafter\ifx\csname natexlab\endcsname\relax\def\natexlab#1{#1}\fi

\bibitem[{Abu-Jbara et~al.(2013)Abu-Jbara, Ezra, and
  Radev}]{abu-jbara-etal-2013-purpose}
Amjad Abu-Jbara, Jefferson Ezra, and Dragomir Radev. 2013.
\newblock \href {https://aclanthology.org/N13-1067} {Purpose and polarity of
  citation: Towards {NLP}-based bibliometrics}.
\newblock In \emph{Proceedings of the 2013 Conference of the North {A}merican
  Chapter of the Association for Computational Linguistics: Human Language
  Technologies}, pages 596--606, Atlanta, Georgia. Association for
  Computational Linguistics.

\bibitem[{Aghion and Howitt(1990)}]{aghion1990model}
Philippe Aghion and Peter Howitt. 1990.
\newblock A model of growth through creative destruction.

\bibitem[{Altafini(2012)}]{altafini2012consensus}
Claudio Altafini. 2012.
\newblock Consensus problems on networks with antagonistic interactions.
\newblock \emph{IEEE transactions on automatic control}, 58(4):935--946.

\bibitem[{Andrade(2011)}]{Andrade2011HowTW}
Chittaranjan Andrade. 2011.
\newblock How to write a good abstract for a scientific paper or conference
  presentation.
\newblock \emph{Indian Journal of Psychiatry}, 53:172 -- 175.

\bibitem[{Athar and Teufel(2012)}]{athar-teufel-2012-context}
Awais Athar and Simone Teufel. 2012.
\newblock \href {https://aclanthology.org/N12-1073} {Context-enhanced citation
  sentiment detection}.
\newblock In \emph{Proceedings of the 2012 Conference of the North {A}merican
  Chapter of the Association for Computational Linguistics: Human Language
  Technologies}, pages 597--601, Montr{\'e}al, Canada. Association for
  Computational Linguistics.

\bibitem[{Beese(2023)}]{zenodo}
Dominik Beese. 2023.
\newblock \href {https://doi.org/10.5281/zenodo.7590215}
  {{DominikBeese/DidAIGetMoreNegativeRecently: Initial release}}.

\bibitem[{Beltagy et~al.(2019)Beltagy, Lo, and Cohan}]{SciBERT}
Iz~Beltagy, Kyle Lo, and Arman Cohan. 2019.
\newblock \href {https://doi.org/10.18653/v1/D19-1371} {{S}ci{BERT}: A
  pretrained language model for scientific text}.
\newblock In \emph{Proceedings of the 2019 Conference on Empirical Methods in
  Natural Language Processing and the 9th International Joint Conference on
  Natural Language Processing (EMNLP-IJCNLP)}, pages 3615--3620, Hong Kong,
  China. Association for Computational Linguistics.

\bibitem[{Bojar et~al.(2010)Bojar, Kos, and
  Mare{\v{c}}ek}]{bojar-etal-2010-tackling}
Ond{\v{r}}ej Bojar, Kamil Kos, and David Mare{\v{c}}ek. 2010.
\newblock \href {https://aclanthology.org/P10-2016} {Tackling sparse data issue
  in machine translation evaluation}.
\newblock In \emph{Proceedings of the {ACL} 2010 Conference Short Papers},
  pages 86--91, Uppsala, Sweden. Association for Computational Linguistics.

\bibitem[{Bordignon(2020)}]{bordignon2020self}
Frederique Bordignon. 2020.
\newblock Self-correction of science: a comparative study of negative citations
  and post-publication peer review.
\newblock \emph{Scientometrics}, 124:1225--1239.

\bibitem[{Borthen(2004)}]{borthen-2004-predicative}
Kaja Borthen. 2004.
\newblock \href {https://aclanthology.org/C04-1169} {Predicative {NP}s and the
  annotation of reference chains}.
\newblock In \emph{{COLING} 2004: Proceedings of the 20th International
  Conference on Computational Linguistics}, pages 1175--1178, Geneva,
  Switzerland. COLING.

\bibitem[{Bowman(2022)}]{bowman-2022-dangers}
Samuel Bowman. 2022.
\newblock \href {https://aclanthology.org/2022.acl-long.516} {The dangers of
  underclaiming: Reasons for caution when reporting how {NLP} systems fail}.
\newblock In \emph{Proceedings of the 60th Annual Meeting of the Association
  for Computational Linguistics (Volume 1: Long Papers)}, pages 7484--7499,
  Dublin, Ireland. Association for Computational Linguistics.

\bibitem[{Catalini et~al.(2015)Catalini, Lacetera, and Oettl}]{catalini:2015}
Christian Catalini, Nicola Lacetera, and Alexander Oettl. 2015.
\newblock \href {https://doi.org/10.1073/pnas.1502280112} {The incidence and
  role of negative citations in science}.
\newblock \emph{Proceedings of the National Academy of Sciences},
  112(45):13823--13826.

\bibitem[{Chen et~al.(2022)Chen, Belouadi, and
  Eger}]{Chen2022ReproducibilityIF}
Yanran Chen, Jonas Belouadi, and Steffen Eger. 2022.
\newblock Reproducibility issues for bert-based evaluation metrics.
\newblock In \emph{Conference on Empirical Methods in Natural Language
  Processing}, volume abs/2204.00004.

\bibitem[{Chicco et~al.(2021)Chicco, Warrens, and Jurman}]{R2IsBetterThanMSE}
Davide Chicco, Matthijs~J Warrens, and Giuseppe Jurman. 2021.
\newblock The coefficient of determination r-squared is more informative than
  smape, mae, mape, mse and rmse in regression analysis evaluation.
\newblock \emph{PeerJ Computer Science}, 7:e623.

\bibitem[{Cohan et~al.(2019)Cohan, Ammar, van Zuylen, and
  Cady}]{cohan-etal-2019-structural}
Arman Cohan, Waleed Ammar, Madeleine van Zuylen, and Field Cady. 2019.
\newblock \href {https://doi.org/10.18653/v1/N19-1361} {Structural scaffolds
  for citation intent classification in scientific publications}.
\newblock In \emph{Proceedings of the 2019 Conference of the North {A}merican
  Chapter of the Association for Computational Linguistics: Human Language
  Technologies, Volume 1 (Long and Short Papers)}, pages 3586--3596,
  Minneapolis, Minnesota. Association for Computational Linguistics.

\bibitem[{Collins(1994)}]{Collins1994}
Randall Collins. 1994.
\newblock \href {https://doi.org/10.1007/BF01476360} {Why the social sciences
  won't become high-consensus, rapid-discovery science}.
\newblock \emph{Sociological Forum}, 9(2):155--177.

\bibitem[{Corkery et~al.(2019)Corkery, Matusevych, and
  Goldwater}]{corkery-etal-2019-yet}
Maria Corkery, Yevgen Matusevych, and Sharon Goldwater. 2019.
\newblock \href {https://doi.org/10.18653/v1/P19-1376} {Are we there yet?
  encoder-decoder neural networks as cognitive models of {E}nglish past tense
  inflection}.
\newblock In \emph{Proceedings of the 57th Annual Meeting of the Association
  for Computational Linguistics}, pages 3868--3877, Florence, Italy.
  Association for Computational Linguistics.

\bibitem[{Devlin et~al.(2018)Devlin, Chang, Lee, and
  Toutanova}]{devlin2018bert}
Jacob Devlin, Ming-Wei Chang, Kenton Lee, and Kristina Toutanova. 2018.
\newblock Bert: Pre-training of deep bidirectional transformers for language
  understanding.
\newblock \emph{arXiv preprint arXiv:1810.04805}.

\bibitem[{Ding and Palmer(2005)}]{ding-palmer-2005-machine}
Yuan Ding and Martha Palmer. 2005.
\newblock \href {https://doi.org/10.3115/1219840.1219907} {Machine translation
  using probabilistic synchronous dependency insertion grammars}.
\newblock In \emph{Proceedings of the 43rd Annual Meeting of the Association
  for Computational Linguistics ({ACL}{'}05)}, pages 541--548, Ann Arbor,
  Michigan. Association for Computational Linguistics.

\bibitem[{Dunning(1993)}]{dunning-1993-accurate}
Ted Dunning. 1993.
\newblock \href {https://aclanthology.org/J93-1003} {Accurate methods for the
  statistics of surprise and coincidence}.
\newblock \emph{Computational Linguistics}, 19(1):61--74.

\bibitem[{Eger(2016)}]{eger2016opinion}
Steffen Eger. 2016.
\newblock Opinion dynamics and wisdom under out-group discrimination.
\newblock \emph{Mathematical Social Sciences}, 80:97--107.

\bibitem[{Fass(1991)}]{fass-1991-met}
Dan Fass. 1991.
\newblock \href {https://aclanthology.org/J91-1003} {met*: A method for
  discriminating metonymy and metaphor by computer}.
\newblock \emph{Computational Linguistics}, 17(1):49--90.

\bibitem[{Fortunato et~al.(2018)Fortunato, Bergstrom, B{\"o}rner, Evans,
  Helbing, Milojevi{\'c}, Petersen, Radicchi, Sinatra, Uzzi
  et~al.}]{fortunato2018science}
Santo Fortunato, Carl~T Bergstrom, Katy B{\"o}rner, James~A Evans, Dirk
  Helbing, Sta{\v{s}}a Milojevi{\'c}, Alexander~M Petersen, Filippo Radicchi,
  Roberta Sinatra, Brian Uzzi, et~al. 2018.
\newblock Science of science.
\newblock \emph{Science}, 359(6379).

\bibitem[{Gao et~al.(2019)Gao, Eger, Kuznetsov, Gurevych, and
  Miyao}]{DoesMyRebuttalMatter}
Yang Gao, Steffen Eger, Ilia Kuznetsov, Iryna Gurevych, and Yusuke Miyao. 2019.
\newblock \href {https://doi.org/10.18653/v1/N19-1129} {Does my rebuttal
  matter? insights from a major {NLP} conference}.
\newblock In \emph{Proceedings of the 2019 Conference of the North {A}merican
  Chapter of the Association for Computational Linguistics: Human Language
  Technologies, Volume 1 (Long and Short Papers)}, pages 1274--1290,
  Minneapolis, Minnesota. Association for Computational Linguistics.

\bibitem[{Gella and Keller(2017)}]{gella-keller-2017-analysis}
Spandana Gella and Frank Keller. 2017.
\newblock \href {https://doi.org/10.18653/v1/P17-2011} {An analysis of action
  recognition datasets for language and vision tasks}.
\newblock In \emph{Proceedings of the 55th Annual Meeting of the Association
  for Computational Linguistics (Volume 2: Short Papers)}, pages 64--71,
  Vancouver, Canada. Association for Computational Linguistics.

\bibitem[{Gerlach and Altmann(2014)}]{Gerlach2014ScalingLA}
Martin Gerlach and Eduardo~G. Altmann. 2014.
\newblock Scaling laws and fluctuations in the statistics of word frequencies.
\newblock \emph{New Journal of Physics}, 16.

\bibitem[{Gorman and Bedrick(2019)}]{gorman-bedrick-2019-need}
Kyle Gorman and Steven Bedrick. 2019.
\newblock \href {https://doi.org/10.18653/v1/P19-1267} {We need to talk about
  standard splits}.
\newblock In \emph{Proceedings of the 57th Annual Meeting of the Association
  for Computational Linguistics}, pages 2786--2791, Florence, Italy.
  Association for Computational Linguistics.

\bibitem[{Gunning(1952)}]{Gunning1952TheTO}
Robbie Gunning. 1952.
\newblock \emph{The Technique of Clear Writing.}
\newblock McGraw-Hill.

\bibitem[{Guo et~al.(2020)Guo, Jin, Li, Yang, Xu, Ju, Zhang, Xuan, Liu, Su
  et~al.}]{guo2020application}
Qinghua Guo, Shichao Jin, Min Li, Qiuli Yang, Kexin Xu, Yuanzhen Ju, Jing
  Zhang, Jing Xuan, Jin Liu, Yanjun Su, et~al. 2020.
\newblock Application of deep learning in ecological resource research:
  Theories, methods, and challenges.
\newblock \emph{Science China Earth Sciences}, pages 1--18.

\bibitem[{Gupta et~al.(2021)Gupta, Kvernadze, and
  Srikumar}]{Gupta_Kvernadze_Srikumar_2021}
Ashim Gupta, Giorgi Kvernadze, and Vivek Srikumar. 2021.
\newblock \href {https://ojs.aaai.org/index.php/AAAI/article/view/17531} {Bert
  \& family eat word salad: Experiments with text understanding}.
\newblock \emph{Proceedings of the AAAI Conference on Artificial Intelligence},
  35(14):12946--12954.

\bibitem[{Gupta et~al.(2016)Gupta, Karnick, Bansal, and
  Jhala}]{gupta-etal-2016-product}
Vivek Gupta, Harish Karnick, Ashendra Bansal, and Pradhuman Jhala. 2016.
\newblock \href {https://aclanthology.org/C16-1052} {Product classification in
  {E}-commerce using distributional semantics}.
\newblock In \emph{Proceedings of {COLING} 2016, the 26th International
  Conference on Computational Linguistics: Technical Papers}, pages 536--546,
  Osaka, Japan. The COLING 2016 Organizing Committee.

\bibitem[{Hartmann et~al.(2022)Hartmann, Heitmann, Siebert, and
  Schamp}]{HARTMANN2022}
Jochen Hartmann, Mark Heitmann, Christian Siebert, and Christina Schamp. 2022.
\newblock \href
  {https://doi.org/https://doi.org/10.1016/j.ijresmar.2022.05.005} {More than a
  feeling: Accuracy and application of sentiment analysis}.
\newblock \emph{International Journal of Research in Marketing}.

\bibitem[{Hendler(2008)}]{Hendler:2008}
James Hendler. 2008.
\newblock \href {https://doi.org/10.1109/MIS.2008.20} {Avoiding another ai
  winter}.
\newblock \emph{IEEE Intelligent Systems}, 23(2):2--4.

\bibitem[{Herdan(1960)}]{Herdan1960LanguageAC}
Gustav Herdan. 1960.
\newblock \emph{Language as choice and chance}.
\newblock Springer Berlin Heidelberg.

\bibitem[{Howard and Ruder(2018)}]{ULMFiT}
Jeremy Howard and Sebastian Ruder. 2018.
\newblock \href {https://doi.org/10.18653/v1/P18-1031} {Universal language
  model fine-tuning for text classification}.
\newblock In \emph{Proceedings of the 56th Annual Meeting of the Association
  for Computational Linguistics (Volume 1: Long Papers)}, pages 328--339,
  Melbourne, Australia. Association for Computational Linguistics.

\bibitem[{Hutto and Gilbert(2014)}]{VADER}
C.~Hutto and Eric Gilbert. 2014.
\newblock \href {https://ojs.aaai.org/index.php/ICWSM/article/view/14550}
  {Vader: A parsimonious rule-based model for sentiment analysis of social
  media text}.
\newblock \emph{Proceedings of the International AAAI Conference on Web and
  Social Media}, 8(1):216--225.

\bibitem[{Jebari et~al.(2021)Jebari, Herrera-Viedma, and Cobo}]{jebari2021use}
Chaker Jebari, Enrique Herrera-Viedma, and Manuel~Jesus Cobo. 2021.
\newblock \href {https://doi.org/10.1007/s11192-020-03858-y} {The use of
  citation context to detect the evolution of research topics: a large-scale
  analysis}.
\newblock \emph{Scientometrics}, 126(4):2971--2989.

\bibitem[{Jurgens et~al.(2018)Jurgens, Kumar, Hoover, McFarland, and
  Jurafsky}]{MeasuringTheEvolution}
David Jurgens, Srijan Kumar, Raine Hoover, Dan McFarland, and Dan Jurafsky.
  2018.
\newblock \href {https://doi.org/10.1162/tacl_a_00028} {Measuring the evolution
  of a scientific field through citation frames}.
\newblock \emph{Transactions of the Association for Computational Linguistics},
  6:391--406.

\bibitem[{Kang et~al.(2018)Kang, Ammar, Dalvi, van Zuylen, Kohlmeier, Hovy, and
  Schwartz}]{PeerReviewsDataset}
Dongyeop Kang, Waleed Ammar, Bhavana Dalvi, Madeleine van Zuylen, Sebastian
  Kohlmeier, Eduard Hovy, and Roy Schwartz. 2018.
\newblock \href {https://doi.org/10.18653/v1/N18-1149} {A dataset of peer
  reviews ({P}eer{R}ead): Collection, insights and {NLP} applications}.
\newblock In \emph{Proceedings of the 2018 Conference of the North {A}merican
  Chapter of the Association for Computational Linguistics: Human Language
  Technologies, Volume 1 (Long Papers)}, pages 1647--1661, New Orleans,
  Louisiana. Association for Computational Linguistics.

\bibitem[{Kingma and Ba(2015)}]{Adam}
Diederik~P. Kingma and Jimmy Ba. 2015.
\newblock Adam: A method for stochastic optimization.
\newblock \emph{CoRR}, abs/1412.6980.

\bibitem[{Krizhevsky et~al.(2012)Krizhevsky, Sutskever, and
  Hinton}]{krizhevsky2012imagenet}
Alex Krizhevsky, Ilya Sutskever, and Geoffrey~E Hinton. 2012.
\newblock Imagenet classification with deep convolutional neural networks.
\newblock \emph{Advances in neural information processing systems},
  25:1097--1105.

\bibitem[{Kruskal and Wallis(1952)}]{KruskalWallisTest}
William~H. Kruskal and W.~Allen Wallis. 1952.
\newblock \href {https://doi.org/10.1080/01621459.1952.10483441} {Use of ranks
  in one-criterion variance analysis}.
\newblock \emph{Journal of the American Statistical Association},
  47(260):583--621.

\bibitem[{Kurakin et~al.(2018)Kurakin, Goodfellow, and
  Bengio}]{kurakin2018adversarial}
Alexey Kurakin, Ian~J Goodfellow, and Samy Bengio. 2018.
\newblock Adversarial examples in the physical world.
\newblock In \emph{Artificial intelligence safety and security}, pages 99--112.
  Chapman and Hall/CRC.

\bibitem[{Lamers et~al.(2021)Lamers, Boyack, Larivi{\`e}re, Sugimoto, van Eck,
  Waltman, and Murray}]{lamers2021meta}
Wout~S Lamers, Kevin Boyack, Vincent Larivi{\`e}re, Cassidy~R Sugimoto,
  Nees~Jan van Eck, Ludo Waltman, and Dakota Murray. 2021.
\newblock Meta-research: Investigating disagreement in the scientific
  literature.
\newblock \emph{Elife}, 10:e72737.

\bibitem[{Lample et~al.(2018)Lample, Conneau, Denoyer, and
  Ranzato}]{lample2018unsupervised}
Guillaume Lample, Alexis Conneau, Ludovic Denoyer, and Marc'Aurelio Ranzato.
  2018.
\newblock \href {https://openreview.net/forum?id=rkYTTf-AZ} {Unsupervised
  machine translation using monolingual corpora only}.
\newblock In \emph{International Conference on Learning Representations}.

\bibitem[{Lauscher et~al.(2022)Lauscher, Ko, Kuehl, Johnson, Cohan, Jurgens,
  and Lo}]{lauscher2021multicite}
Anne Lauscher, Brandon Ko, Bailey Kuehl, Sophie Johnson, Arman Cohan, David
  Jurgens, and Kyle Lo. 2022.
\newblock \href {https://doi.org/10.18653/v1/2022.naacl-main.137}
  {{M}ulti{C}ite: Modeling realistic citations requires moving beyond the
  single-sentence single-label setting}.
\newblock In \emph{Proceedings of the 2022 Conference of the North American
  Chapter of the Association for Computational Linguistics: Human Language
  Technologies}, pages 1875--1889, Seattle, United States. Association for
  Computational Linguistics.

\bibitem[{Lazaridou et~al.(2021)Lazaridou, Kuncoro, Gribovskaya, Agrawal,
  Liska, Terzi, Gimenez, de~Masson~d'Autume, Ko{\v{c}}isk{\'y}, Ruder,
  Yogatama, Cao, Young, and Blunsom}]{lazaridou2021pitfalls}
Angeliki Lazaridou, Adhiguna Kuncoro, Elena Gribovskaya, Devang Agrawal, Adam
  Liska, Tayfun Terzi, Mai Gimenez, Cyprien de~Masson~d'Autume, Tom{\'a}{\v{s}}
  Ko{\v{c}}isk{\'y}, Sebastian Ruder, Dani Yogatama, Kris Cao, Susannah Young,
  and Phil Blunsom. 2021.
\newblock \href {https://openreview.net/forum?id=73OmmrCfSyy} {Mind the gap:
  Assessing temporal generalization in neural language models}.
\newblock In \emph{Thirty-Fifth Conference on Neural Information Processing
  Systems}.

\bibitem[{Letchford et~al.(2015)Letchford, Moat, and
  Preis}]{letchford2015advantage}
Adrian Letchford, Helen~Susannah Moat, and Tobias Preis. 2015.
\newblock The advantage of short paper titles.
\newblock \emph{Royal Society open science}, 2(8):150266.

\bibitem[{Liang et~al.(2019)Liang, Du, Xu, Li, and
  Huang}]{liang-etal-2019-context}
Bin Liang, Jiachen Du, Ruifeng Xu, Binyang Li, and Hejiao Huang. 2019.
\newblock \href {https://doi.org/10.18653/v1/P19-1462} {Context-aware embedding
  for targeted aspect-based sentiment analysis}.
\newblock In \emph{Proceedings of the 57th Annual Meeting of the Association
  for Computational Linguistics}, pages 4678--4683, Florence, Italy.
  Association for Computational Linguistics.

\bibitem[{Lipton and Steinhardt(2019)}]{Lipton:2019}
Zachary~C. Lipton and Jacob Steinhardt. 2019.
\newblock \href {https://doi.org/10.1145/3317287.3328534} {Troubling trends in
  machine learning scholarship: Some ml papers suffer from flaws that could
  mislead the public and stymie future research.}
\newblock \emph{Queue}, 17(1):45–77.

\bibitem[{Liu and Liu(2009)}]{liu-liu-2009-extractive}
Fei Liu and Yang Liu. 2009.
\newblock \href {https://aclanthology.org/P09-2066} {From extractive to
  abstractive meeting summaries: Can it be done by sentence compression?}
\newblock In \emph{Proceedings of the {ACL}-{IJCNLP} 2009 Conference Short
  Papers}, pages 261--264, Suntec, Singapore. Association for Computational
  Linguistics.

\bibitem[{Liu et~al.(2019)Liu, Ott, Goyal, Du, Joshi, Chen, Levy, Lewis,
  Zettlemoyer, and Stoyanov}]{RoBERTa}
Yinhan Liu, Myle Ott, Naman Goyal, Jingfei Du, Mandar Joshi, Danqi Chen, Omer
  Levy, Mike Lewis, Luke Zettlemoyer, and Veselin Stoyanov. 2019.
\newblock \href {https://doi.org/10.48550/ARXIV.1907.11692} {Roberta: A
  robustly optimized bert pretraining approach}.

\bibitem[{Marcus(2018)}]{Marcus2018DeepLA}
Gary~F. Marcus. 2018.
\newblock Deep learning: A critical appraisal.
\newblock \emph{ArXiv}, abs/1801.00631.

\bibitem[{Marie et~al.(2021)Marie, Fujita, and
  Rubino}]{marie-etal-2021-scientific}
Benjamin Marie, Atsushi Fujita, and Raphael Rubino. 2021.
\newblock \href {https://doi.org/10.18653/v1/2021.acl-long.566} {Scientific
  credibility of machine translation research: A meta-evaluation of 769
  papers}.
\newblock In \emph{Proceedings of the 59th Annual Meeting of the Association
  for Computational Linguistics and the 11th International Joint Conference on
  Natural Language Processing (Volume 1: Long Papers)}, pages 7297--7306,
  Online. Association for Computational Linguistics.

\bibitem[{Mlinari{\'c} et~al.(2017)Mlinari{\'c}, Horvat, and
  {\v{S}}upak~Smol{\v{c}}i{\'c}}]{mlinaric2017dealing}
Ana Mlinari{\'c}, Martina Horvat, and Vesna {\v{S}}upak~Smol{\v{c}}i{\'c}.
  2017.
\newblock Dealing with the positive publication bias: Why you should really
  publish your negative results.
\newblock \emph{Biochemia medica}, 27(3):447--452.

\bibitem[{Mohammed et~al.(2020)Mohammed, Rawashdeh, and Abdullah}]{9078901}
Roweida Mohammed, Jumanah Rawashdeh, and Malak Abdullah. 2020.
\newblock \href {https://doi.org/10.1109/ICICS49469.2020.239556} {Machine
  learning with oversampling and undersampling techniques: Overview study and
  experimental results}.
\newblock In \emph{2020 11th International Conference on Information and
  Communication Systems (ICICS)}, pages 243--248.

\bibitem[{Niven and Kao(2019)}]{niven-kao-2019-probing}
Timothy Niven and Hung-Yu Kao. 2019.
\newblock \href {https://doi.org/10.18653/v1/P19-1459} {Probing neural network
  comprehension of natural language arguments}.
\newblock In \emph{Proceedings of the 57th Annual Meeting of the Association
  for Computational Linguistics}, pages 4658--4664, Florence, Italy.
  Association for Computational Linguistics.

\bibitem[{Pei and Jurgens(2021)}]{pei-jurgens-2021-measuring}
Jiaxin Pei and David Jurgens. 2021.
\newblock \href {https://doi.org/10.18653/v1/2021.emnlp-main.784} {Measuring
  sentence-level and aspect-level (un)certainty in science communications}.
\newblock In \emph{Proceedings of the 2021 Conference on Empirical Methods in
  Natural Language Processing}, pages 9959--10011, Online and Punta Cana,
  Dominican Republic. Association for Computational Linguistics.

\bibitem[{Post(2018)}]{post-2018-call}
Matt Post. 2018.
\newblock \href {https://doi.org/10.18653/v1/W18-6319} {A call for clarity in
  reporting {BLEU} scores}.
\newblock In \emph{Proceedings of the Third Conference on Machine Translation:
  Research Papers}, pages 186--191, Brussels, Belgium. Association for
  Computational Linguistics.

\bibitem[{Prabhakaran et~al.(2016)Prabhakaran, Hamilton, McFarland, and
  Jurafsky}]{PredictingTheRiseAndFall}
Vinodkumar Prabhakaran, William~L. Hamilton, Dan McFarland, and Dan Jurafsky.
  2016.
\newblock \href {https://doi.org/10.18653/v1/P16-1111} {Predicting the rise and
  fall of scientific topics from trends in their rhetorical framing}.
\newblock In \emph{Proceedings of the 54th Annual Meeting of the Association
  for Computational Linguistics (Volume 1: Long Papers)}, pages 1170--1180,
  Berlin, Germany. Association for Computational Linguistics.

\bibitem[{Radford et~al.(2019)Radford, Wu, Child, Luan, Amodei, Sutskever
  et~al.}]{radford2019language}
Alec Radford, Jeffrey Wu, Rewon Child, David Luan, Dario Amodei, Ilya
  Sutskever, et~al. 2019.
\newblock Language models are unsupervised multitask learners.
\newblock \emph{OpenAI blog}, 1(8):9.

\bibitem[{Reimers and Gurevych(2017)}]{Reimers2017ReportingSD}
Nils Reimers and Iryna Gurevych. 2017.
\newblock Reporting score distributions makes a difference: Performance study
  of lstm-networks for sequence tagging.
\newblock In \emph{Conference on Empirical Methods in Natural Language
  Processing}.

\bibitem[{Samangouei et~al.(2018)Samangouei, Kabkab, and
  Chellappa}]{samangouei2018defensegan}
Pouya Samangouei, Maya Kabkab, and Rama Chellappa. 2018.
\newblock \href {https://openreview.net/forum?id=BkJ3ibb0-} {Defense-{GAN}:
  Protecting classifiers against adversarial attacks using generative models}.
\newblock In \emph{International Conference on Learning Representations}.

\bibitem[{Sienkiewicz and Altmann(2016)}]{sienkiewicz2016impact}
Julian Sienkiewicz and Eduardo~G Altmann. 2016.
\newblock Impact of lexical and sentiment factors on the popularity of
  scientific papers.
\newblock \emph{Royal Society open science}, 3(6):160140.

\bibitem[{S{\o}gaard et~al.(2018)S{\o}gaard, Ruder, and
  Vuli{\'c}}]{sogaard-etal-2018-limitations}
Anders S{\o}gaard, Sebastian Ruder, and Ivan Vuli{\'c}. 2018.
\newblock \href {https://doi.org/10.18653/v1/P18-1072} {On the limitations of
  unsupervised bilingual dictionary induction}.
\newblock In \emph{Proceedings of the 56th Annual Meeting of the Association
  for Computational Linguistics (Volume 1: Long Papers)}, pages 778--788,
  Melbourne, Australia. Association for Computational Linguistics.

\bibitem[{Teufel et~al.(2006)Teufel, Siddharthan, and
  Tidhar}]{teufel_automatic}
Simone Teufel, Advaith Siddharthan, and Dan Tidhar. 2006.
\newblock Automatic classification of citation function.
\newblock In \emph{Proceedings of the 2006 Conference on Empirical Methods in
  Natural Language Processing}, EMNLP '06, page 103–110, USA. Association for
  Computational Linguistics.

\bibitem[{Wang et~al.(2020)Wang, Zeng, Huang, Knight, Ji, and
  Rajani}]{ReviewRobot}
Qingyun Wang, Qi~Zeng, Lifu Huang, Kevin Knight, Heng Ji, and Nazneen~Fatema
  Rajani. 2020.
\newblock \href {https://aclanthology.org/2020.inlg-1.44} {{R}eview{R}obot:
  Explainable paper review generation based on knowledge synthesis}.
\newblock In \emph{Proceedings of the 13th International Conference on Natural
  Language Generation}, pages 384--397, Dublin, Ireland. Association for
  Computational Linguistics.

\bibitem[{Warriner et~al.(2013)Warriner, Kuperman, and
  Brysbaert}]{Warriner2013-vw}
Amy~Beth Warriner, Victor Kuperman, and Marc Brysbaert. 2013.
\newblock Norms of valence, arousal, and dominance for 13,915 english lemmas.
\newblock \emph{Behavior Research Methods}, 45(4):1191--1207.

\bibitem[{Welch(1947)}]{WelchTest}
B.~L. Welch. 1947.
\newblock \href {https://doi.org/10.1093/biomet/34.1-2.28} {{The generalisation
  of student's problems when several different population variances are
  involved}}.
\newblock \emph{Biometrika}, 34(1-2):28--35.

\bibitem[{Wright and Augenstein(2021)}]{wright-augenstein-2021-citeworth}
Dustin Wright and Isabelle Augenstein. 2021.
\newblock \href {https://doi.org/10.18653/v1/2021.findings-acl.157}
  {{C}ite{W}orth: Cite-worthiness detection for improved scientific document
  understanding}.
\newblock In \emph{Findings of the Association for Computational Linguistics:
  ACL-IJCNLP 2021}, pages 1796--1807, Online. Association for Computational
  Linguistics.

\bibitem[{Wu et~al.(2018)Wu, Ren, Liao, and Grosse.}]{wu2018understanding}
Yuhuai Wu, Mengye Ren, Renjie Liao, and Roger Grosse. 2018.
\newblock \href {https://openreview.net/forum?id=H1MczcgR-} {Understanding
  short-horizon bias in stochastic meta-optimization}.
\newblock In \emph{International Conference on Learning Representations}.

\bibitem[{Yan et~al.(2011)Yan, Tang, Liu, Shan, and Li}]{yan2011citation}
Rui Yan, Jie Tang, Xiaobing Liu, Dongdong Shan, and Xiaoming Li. 2011.
\newblock Citation count prediction: learning to estimate future citations for
  literature.
\newblock In \emph{Proceedings of the 20th ACM international conference on
  Information and knowledge management}, pages 1247--1252.

\bibitem[{Yousif et~al.(2019)Yousif, Niu, Tarus, and Ahmad}]{yousif2019survey}
Abdallah Yousif, Zhendong Niu, John~K Tarus, and Arshad Ahmad. 2019.
\newblock A survey on sentiment analysis of scientific citations.
\newblock \emph{Artificial Intelligence Review}, 52(3):1805--1838.

\bibitem[{Yuan et~al.(2021)Yuan, Liu, and Neubig}]{CanWeAutomate}
Weizhe Yuan, Pengfei Liu, and Graham Neubig. 2021.
\newblock \href {http://arxiv.org/abs/2102.00176} {Can we automate scientific
  reviewing?}
\newblock \emph{CoRR}, abs/2102.00176.

\end{thebibliography}

\renewcommand{\thesection}{\Alph{section}}
\setcounter{section}{0}
\section{Appendix}\customlabel{sec:appendix}{Appendix}

\subsection{Annotation guidelines}
We issued the following guidelines to annotators:
(i) the stance is a value in the range from $-$1 (very negative) to +1 (very positive), and
(ii) only the title and abstract of a paper is taken into account.

A contribution has a \emph{positive stance} and is annotated with a positive number up to +1 when:
(i) it clearly indicates to improve the state-of-the-art by beating existing standards;
(ii) it presents novel techniques;
(iii) it proposes solutions to problems of previous work;
(iv) it gives insights to existing models or methods and explains why they work.

A contribution has a \emph{negative stance} and is annotated with a negative number up to $-$1 when:
(i) it clearly criticizes previous work for being wrong;
(ii) it presents flaws of existing work, i.e. that an approach is deficient with respect to some property;
(iii) it analyses errors of other methods and explains why they do not work as expected.

Contributions that have positive and negative parts are annotated with a value between $-$1 and +1, taking into account the following:
(i) the importance of individual parts matters, i.e. `some problems' is less negative than `fails to work';
(ii) the amount of positive and negative parts matters;
(iii) the last sentence of an abstract is usually the most important sentence. If the last sentence of a contribution is positive, it is more positive than a contribution with a negative last sentence.

Contributions that fall outside this labelling scheme are \emph{neutral} and annotated with a 0. Those include:
(i) contributions that explore existing work without beating other systems or explaining why it works or does not work;
(ii) contributions that compare, discuss, study or summarize existing work without criticizing it.

\subsection{Range definitions}
We show that our trends are similar with different thresholds for positive (greater than or equal to $0.1$), negative (less than or equal to $-0.1$), and neutral ($\in (-0.1,+0.1)$) papers by recreating \ref{fig:negative_stance_per_year_and_domain} and \ref{fig:negative_stance_per_year_and_venue} with alternative thresholds for positive (greater than or equal to $0.2$), negative (less than or equal to $-0.2$), and neutral ($\in (-0.2,+0.2)$) papers, cf.\ \ref{fig:negative_stance_per_year_and_domain_0.2} and \ref{fig:negative_stance_per_year_and_venue_0.2}.

\begin{figure}[!htb]
    \centering
    \includegraphics[width=\linewidth]{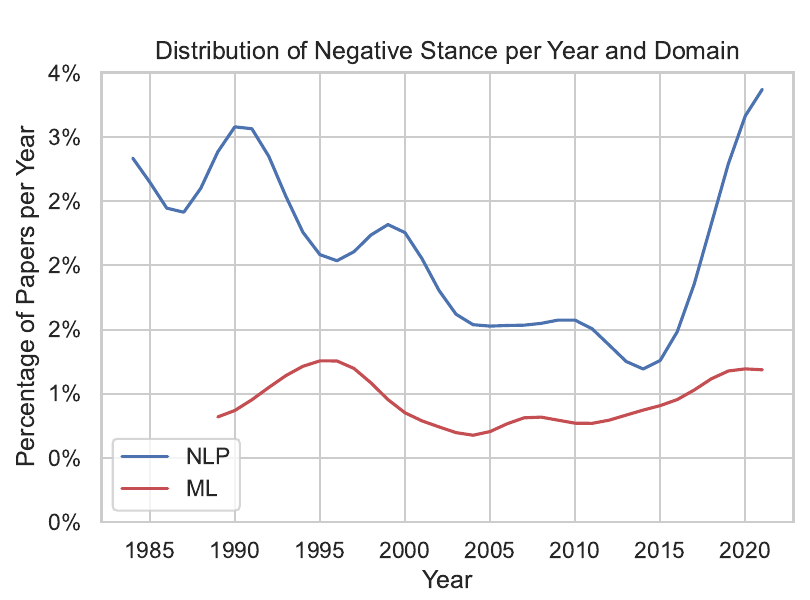}
    \caption{Percentage of papers with a negative stance, calculated as the number of negative papers divided by the total number of papers in each year, for both domains with \emph{alternative thresholds}.}
    \figurelabel{fig:negative_stance_per_year_and_domain_0.2}
\end{figure}

\begin{figure}[!htb]
    \centering
    \includegraphics[width=\linewidth]{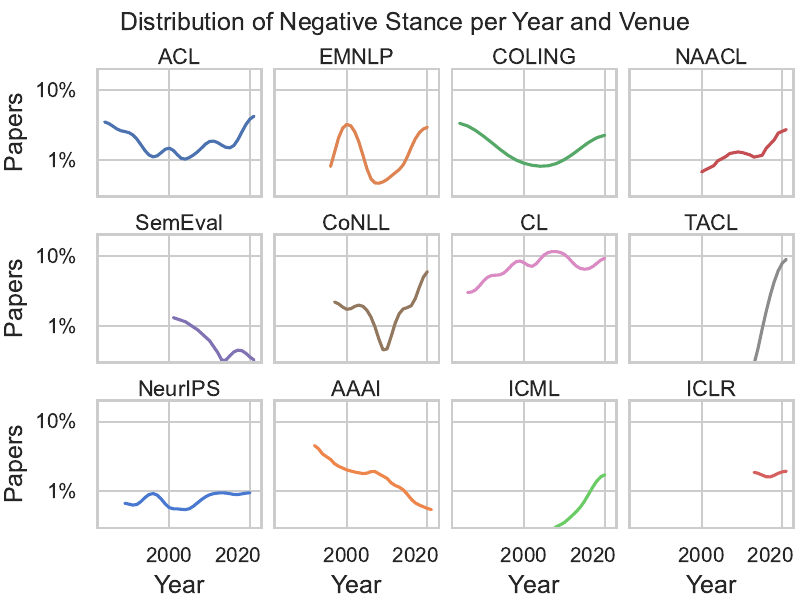}
    \caption{Percentage of negative papers, calculated as the number of negative papers divided by the total number of papers in each year, for each venue on a logarithmic scale with \emph{alternative thresholds}.}
    \figurelabel{fig:negative_stance_per_year_and_venue_0.2}
\end{figure}

\begin{table*}[!htb]
    \centering
    {\footnotesize\begin{tabular}{l|l|l}
        \toprule
        \textbf{Positive n-grams} & \textbf{Neutral n-grams} & \textbf{Negative n-grams} \\
        \midrule
        we propose                       & computational linguistics & we find                \\
        propose a                        & of the                    & that these             \\
        we propose a                     & the language              & and that               \\
        our approach                     & to be                     & find that              \\
        our method                       & university of             & conclude that          \\
        that our                         & of course                 & fails to               \\
        our model                        & in the                    & neural models          \\
        a novel                          & in fact                   & evidence that          \\
        the proposed                     & the following             & we find that           \\
        show that our                    & is not                    & argue that             \\
        paper we propose                 & is a                      & our findings           \\
        this paper we propose            & it is                     & fail to                \\
        we present                       & of it                     & adversarial examples   \\
        we present a                     & there is                  & we show that           \\
        experimental results             & question of               & that the               \\
        of our                           & there are                 & do not                 \\
        propose a novel                  & of some                   & results suggest that   \\
        demonstrate the                  & the formal                & we analyze             \\
        we propose a novel               & about the                 & suggest that           \\
        we introduce                     & try to                    & language understanding \\
        method for                       & the question of           & that this              \\
        present a                        & of natural language       & the difficulty         \\
        propose an                       & of a                      & is not                 \\
        experiments on                   & number of                 & may not                \\
        an efficient                     & the book                  & to understand          \\
        in this paper                    & there is a                & it has been            \\
        we propose an                    & one can                   & our results            \\
        effectiveness of                 & is no                     & we conclude that       \\
        paper we propose a               & the importance of         & that there is          \\
        method to                        & the role of               & our results suggest    \\
        proposed method                  & properties of the         & for this task          \\
        that our method                  & is not the                & the ability of         \\
        in this paper we                 & need to be                & argue that the         \\
        the effectiveness of             & it is not                 & we prove that          \\
        demonstrate that our             & analysis of the           & we investigate the     \\
        effectiveness of our             & there is no               & the case of            \\
        algorithm for                    & a number of               & performance EOS EOS    \\
        that our approach                & the use of                & that it is             \\
        the effectiveness of our         & a large number            & a series of            \\
        experimental results on          & a large number of         & the lack of            \\
        of the proposed                  & the fact that             & in recent years        \\
        paper we present a               & part of the               & learning EOS EOS       \\
        that our model                   & of the system             & we argue that          \\
        the proposed method              & some of the               & with respect to        \\
        present a novel                  & large number of           & we show that the       \\
        \bottomrule
    \end{tabular}}
    \caption{Top n-grams in the abstract of positive, neutral or negative papers in comparison with abstracts of \mbox{non-positive}, non-neutral or non-negative papers, ranked according to the log-likelihood ratio \citep{dunning-1993-accurate} using the tool from \citet{DoesMyRebuttalMatter}. We use the 1.5 k papers from our human-annotated dataset. EOS denotes the end of sentence.}
    \tablelabel{tab:llr}
\end{table*}

\begin{table*}[!htb]
    \centering
    {\footnotesize
    \begin{tabularx}{\textwidth}{lX}
        \toprule
        \textbf{Stance} & \textbf{Abstract} \\
        \midrule
        \NLP{} (1.0) & \enquote{Syntax-based statistical machine translation (MT) aims at applying statistical models to structured data. In this paper, \positive{we present a syntax-based statistical machine translation system based on a probabilistic synchronous dependency insertion grammar}. Synchronous dependency insertion grammars are a version of synchronous grammars defined on dependency trees. \positive{We first introduce our approach to inducing such a grammar from parallel corpora. Second, we describe the graphical model for the machine translation task, which can also be viewed as a stochastic tree-to-tree transducer. We introduce a polynomial time decoding algorithm for the model.} We evaluate the outputs of our MT system using the NIST and Bleu automatic MT evaluation software. \positive{The result shows that our system outperforms the baseline system based on the IBM models in both translation speed and quality.}} \citep{ding-palmer-2005-machine} \\
        \midrule
        \NLP{} (1.0) & \enquote{Product classification is the task of automatically predicting a taxonomy path for a product in a predefined taxonomy hierarchy given a textual product description or title. For efficient product classification we require a suitable representation for a document (the textual description of a product) feature vector and efficient and fast algorithms for prediction. \positive{To address the above challenges, we propose a new distributional semantics representation for document vector formation. We also develop a new two-level ensemble approach utilising (with respect to the taxonomy tree) path-wise, node-wise and depth-wise classifiers to reduce error in the final product classification task. Our experiments show the effectiveness of the distributional representation and the ensemble approach on data sets from a leading e-commerce platform and achieve improved results on various evaluation metrics compared to earlier approaches.}} \citep{gupta-etal-2016-product} \\
        \midrule
        \NLP{} (0.5) & \enquote{\negative{We illustrate and explain problems of n-grams-based machine translation (MT) metrics (e.g. BLEU) when applied to morphologically rich languages such as Czech.} \positive{A novel metric SemPOS based on the deep-syntactic representation of the sentence tackles the issue and retains the performance for translation to English as well.}} \citep{bojar-etal-2010-tackling} \\
        \midrule
        \NLP{} (0.3) & \enquote{Most previous studies on meeting summarization have focused on extractive summarization. In this paper, we investigate if we can apply sentence compression to extractive summaries to generate abstractive summaries. \positive{We use different compression algorithms, including integer linear programming with an additional step of filler phrase detection, a noisy-channel approach using Markovization formulation of grammar rules, as well as human compressed sentences. Our experiments on the ICSI meeting corpus show that when compared to the abstractive summaries, using sentence compression on the extractive summaries improves their ROUGE scores;} \negative{however, the best performance is still quite low, suggesting the need of language generation for abstractive summarization.}} \citep{liu-liu-2009-extractive} \\
        \midrule
        \NLP{} (0.0) & \enquote{A large amount of recent research has focused on tasks that combine language and vision, resulting in a proliferation of datasets and methods. One such task is action recognition, whose applications include image annotation, scene understanding and image retrieval. In this survey, we categorize the existing approaches based on how they conceptualize this problem and provide a detailed review of existing datasets, highlighting their diversity as well as advantages and disadvantages. We focus on recently developed datasets which link visual information with linguistic resources and provide a fine-grained syntactic and semantic analysis of actions in images.} \citep{gella-keller-2017-analysis} \\
        \bottomrule
    \end{tabularx}}
    \caption{Selected abstracts from our human annotated dataset. Blue/red phrases denote positive and negative text and are highlighted by us \textit{post hoc}. (First part.)}
    \tablelabel{tab:human_data}
\end{table*}

\begin{table*}[!htb]
    \centering
    {\footnotesize
    \begin{tabularx}{\textwidth}{lX}
        \toprule
        \textbf{Stance} & \textbf{Abstract} \\
        \midrule
        \Hist{} \NLP{} (0.0) & \enquote{The met* method distinguishes selected examples of metonymy from metaphor and from literalness and anomaly in short English sentences. In the met* method, literalness is distinguished because it satisfies contextual constraints that the nonliteral others all violate. Metonymy is discriminated from metaphor and anomaly in a way that [1] supports Lakoff and Johnson's (1980) view that in metonymy one entity stands for another whereas in metaphor one entity is viewed as another, [2] permits chains of metonymies (Reddy 1979), and [3] allows metonymies to co-occur with instances of either literalness, metaphor, or anomaly. Metaphor is distinguished from anomaly because the former contains a relevant analogy, unlike the latter. The met* method is part of Collative Semantics, a semantics for natural language processing, and has been implemented in a computer program called meta5. Some examples of meta5's analysis of metaphor and metonymy are given. The met* method is compared with approaches from artificial intelligence, linguistics, philosophy, and psychology.} \citep{fass-1991-met} \\
        \midrule
        \ML{} ($-$0.5) & \enquote{In this paper, we study the response of large models from the BERT family to incoherent inputs that should confuse any model that claims to understand natural language. We define simple heuristics to construct such examples. \negative{Our experiments show that state-of-the-art models consistently fail to recognize them as ill-formed, and instead produce high confidence predictions on them. As a consequence of this phenomenon, models trained on sentences with randomly permuted word order perform close to state-of-the-art models.} \positive{To alleviate these issues, we show that if models are explicitly trained to recognize invalid inputs, they can be robust to such attacks without a drop in performance.}} \citep{Gupta_Kvernadze_Srikumar_2021} \\
        \midrule
        \NLP{} ($-$0.6) & \enquote{Unsupervised machine translation - i.e. not assuming any cross-lingual supervision signal, whether a dictionary, translations, or comparable corpora - seems impossible, but nevertheless, Lample et al. (2017) recently proposed a fully unsupervised machine translation (MT) model. The model relies heavily on an adversarial, unsupervised cross-lingual word embedding technique for bilingual dictionary induction (Conneau et al., 2017), which we examine here. \negative{Our results identify the limitations of current unsupervised MT: unsupervised bilingual dictionary induction performs much worse on morphologically rich languages that are not dependent marking, when monolingual corpora from different domains or different embedding algorithms are used.} \positive{We show that a simple trick, exploiting a weak supervision signal from identical words, enables more robust induction and establish a near-perfect correlation between unsupervised bilingual dictionary induction performance and a previously unexplored graph similarity metric.}} \citep{sogaard-etal-2018-limitations} \\
        \midrule
        \NLP{} ($-$1.0) & \enquote{It is standard practice in speech \& language technology to rank systems according to their performance on a test set held out for evaluation. \negative{However, few researchers apply statistical tests to determine whether differences in performance are likely to arise by chance, and few examine the stability of system ranking across multiple training-testing splits. We conduct replication and reproduction experiments with nine part-of-speech taggers published between 2000 and 2018, each of which claimed state-of-the-art performance on a widely-used \enquote{standard split}. While we replicate results on the standard split, we fail to reliably reproduce some rankings when we repeat this analysis with randomly generated training-testing splits.} We argue that randomly generated splits should be used in system evaluation.} \citep{gorman-bedrick-2019-need} \\
        \midrule 
        \NLP{} ($-$1.0) & \enquote{The cognitive mechanisms needed to account for the English past tense have long been a subject of debate in linguistics and cognitive science. Neural network models were proposed early on, but were shown to have clear flaws. Recently, however, Kirov and Cotterell (2018) showed that modern encoder-decoder (ED) models overcome many of these flaws. They also presented evidence that ED models demonstrate humanlike performance in a nonce-word task. Here, we look more closely at the behaviour of their model in this task. \negative{We find that (1) the model exhibits instability across multiple simulations in terms of its correlation with human data, and (2) even when results are aggregated across simulations (treating each simulation as an individual human participant), the fit to the human data is not strong --- worse than an older rule-based model. These findings hold up through several alternative training regimes and evaluation measures.} Although other neural architectures might do better, \negative{we conclude that there is still insufficient evidence to claim that neural nets are a good cognitive model for this task}.} \citep{corkery-etal-2019-yet} \\
        \bottomrule
    \end{tabularx}}
    \caption{Selected abstracts from our human annotated dataset. Blue/red phrases denote positive and negative text and are highlighted by us \textit{post hoc}. (Second part.)}
    \tablelabel{tab:human_data2}
\end{table*}

\begin{table*}[!htb]
    \centering
    {\footnotesize
    \begin{tabularx}{\textwidth}{lX}
        \toprule
        \textbf{Stance} & \textbf{Abstract} \\
        \midrule
        \NLP{} (0.93) & \enquote{Attention-based neural models were employed to detect the different aspects and sentiment polarities of the same target in targeted aspect-based sentiment analysis (TABSA). \negative{However, existing methods do not specifically pre-train reasonable embeddings for targets and aspects in TABSA. This may result in targets or aspects having the same vector representations in different contexts and losing the context-dependent information.} \positive{To address this problem, we propose a novel method to refine the embeddings of targets and aspects. Such pivotal embedding refinement utilizes a sparse coefficient vector to adjust the embeddings of target and aspect from the context. Hence the embeddings of targets and aspects can be refined from the highly correlative words instead of using context-independent or randomly initialized vectors. Experiment results on two benchmark datasets show that our approach yields the state-of-the-art performance in TABSA task.}} \citep{liang-etal-2019-context} \\
        \midrule
        \ML{} (0.85) & \enquote{In recent years, deep neural network approaches have been widely adopted for machine learning tasks, including classification. \negative{However, they were shown to be vulnerable to adversarial perturbations: carefully crafted small perturbations can cause misclassification of legitimate images.} \positive{We propose Defense-GAN, a new framework leveraging the expressive capability of generative models to defend deep neural networks against such attacks.} Defense-GAN is trained to model the distribution of unperturbed images. At inference time, it finds a close output to a given image which does not contain the adversarial changes. This output is then fed to the classifier. \positive{Our proposed method can be used with any classification model and does not modify the classifier structure or training procedure. It can also be used as a defense against any attack as it does not assume knowledge of the process for generating the adversarial examples. We empirically show that Defense-GAN is consistently effective against different attack methods and improves on existing defense strategies.}} \citep{samangouei2018defensegan} \\
        \midrule
        \NLP{} ($-$0.77) & \enquote{In the development of machine learning systems for identification of reference chains, hand-annotated corpora play a crucial role. This paper concerns the question of how predicative NPs should be annotated w.r.t. coreference in corpora for such systems. This question highlights the tension that sometimes appears in the development of corpora between linguistic considerations and the aim for perfection on the one hand and practical applications and the aim for efficiency on the other. \negative{Many current projects that seek to identify coreferential links automatically, assume an annotation strategy which instructs the annotator to mark a predicative NP as coreferential with its subject if it is part of a positive sentence. This paper argues that such a representation is not linguistically plausible, and that it will fail to generate an optimal result.}} \citep{borthen-2004-predicative} \\
        \midrule
        \ML{} ($-$0.94) & \enquote{Careful tuning of the learning rate, or even schedules thereof, can be crucial to effective neural net training. There has been much recent interest in gradient-based meta-optimization, where one tunes hyperparameters, or even learns an optimizer, in order to minimize the expected loss when the training procedure is unrolled. But because the training procedure must be unrolled thousands of times, the metaobjective must be defined with an orders-of-magnitude shorter time horizon than is typical for neural net training. \negative{We show that such short-horizon meta-objectives cause a serious bias towards small step sizes, an effect we term short-horizon bias.} We introduce a toy problem, a noisy quadratic cost function, on which we analyze short-horizon bias by deriving and comparing the optimal schedules for short and long time horizons. \negative{We then run meta-optimization experiments (both offline and online) on standard benchmark datasets, showing that meta-optimization chooses too small a learning rate by multiple orders of magnitude, even when run with a moderately long time horizon (100 steps) typical of work in the area. We believe short-horizon bias is a fundamental problem that needs to be addressed if metaoptimization is to scale to practical neural net training regimes.}} \citep{wu2018understanding} \\
        \bottomrule
    \end{tabularx}}
    \caption{Predicted highly positive and highly negative papers from our datasets. Blue/red text represents positive/negative rationales; these are added by us.}
    \tablelabel{tab:papers}
\end{table*}

\end{document}